\documentclass[utf8]{FrontiersinHarvard} 

\usepackage{url,hyperref,microtype,subcaption}
\usepackage[onehalfspacing]{setspace}

\usepackage{algorithm}
\usepackage[noend]{algpseudocode}
\usepackage{csquotes}  
\usepackage{color}
\usepackage[normalem]{ulem}
\usepackage{cleveref}
\usepackage{algorithm}
\usepackage{algpseudocode}

\crefname{figure}{Fig.}{Figs.}
\crefname{table}{Tab.}{Tabs.}

\usepackage{todonotes}

\newcommand \reply[1]{\bgroup\noindent[\textcolor{blue}{\textbf{Reply}: #1}]\egroup\ignorespacesafterend}
\newcommand \improve[1]{\bgroup\noindent[\textcolor{green}{\textbf{improve}: #1}]\egroup\ignorespacesafterend} 

\usepackage{natbib}
\setcitestyle{square,numbers,sort&compress,comma}

\DeclareCaptionFont{9pt}{\fontsize{9pt}{10pt}\selectfont}
\usepackage{float}
\usepackage{graphicx}
\usepackage{subcaption}

\def\keyFont{\fontsize{8}{11}\helveticabold }
\def\firstAuthorLast{} 
\def\Authors{Kishan Govind\,$^{1}$,  Daniela Oliveros\,$^{2}$, Antonin Dlouhy\,$^{3}$,Marc Legros\,$^{2}$, and Stefan Sandfeld\,$^{1,4,*}$ }
\def\Address{$^{1}$ Institute for Advanced Simulations: Materials Data Science and Informatics (IAS-9), Forschungszentrum Jülich GmbH, 52425 Jülich, Germany \\
	$^{2}$CEMES-CNRS 29 Rue J. Marvig Toulouse 31055 France \\
	$^{3}$Institute  of Physics  of Materials, Czech Academy  of  Sciences, 61662 Brno Czech Republic \\
	$^{4}$ Faculty of Georesources and Materials Engineering, RWTH Aachen University, 52068 Aachen, Germany}
\usepackage[nolist, smaller]{acronym}
\def\corrEmail{s.sandfeld@fz-juelich.de}
\begin{acronym}
	\acro{DL}[DL]{deep learning}
	\acro{DS}[DS]{dataset}  
	\acro{IoU}[IoU]{Intersection over Union}    
	\acro{ML}[ML]{Machine Learning}
	\acro{MSE}[$\textrm{MSE}$]{mean squared error}
	\acro{TEM}[TEM]{Transmission Electron Microscopy}
	\acro{GAN}[GAN]{Generative Adversial Network}
\end{acronym}

\newcommand{\TEM}{\ac{TEM}}
\newcommand{\DL}{\ac{DL}}

\begin{document}
	\onecolumn
	\firstpage{1}
	
	\title[]{Deep Learning of Crystalline Defects from TEM images: A Solution for the Problem of \enquote{Never Enough Training Data}} 
	
	\author[\firstAuthorLast ]{\Authors} 
	\address{} 
	\correspondance{} 
	
	\extraAuth{}

	\maketitle

\begin{abstract}\small
Crystalline defects, such as line-like dislocations,  play an important role 
for the performance and reliability of many metallic devices. Their 
interaction and evolution still poses a multitude of open questions to materials 
science and materials physics. In-situ TEM experiments can provide important 
insights into how dislocations behave and move. During such experiments, the 
dislocation microstructure is captured in form of videos. The analysis of 
individual video frames can provide useful insights but is limited by the 
capabilities of automated identification, digitization, and quantitative extraction 
of the dislocations as curved objects. The vast amount of data
also makes manual annotation very time consuming, thereby limiting the use of 
Deep Learning-based, automated image analysis and segmentation of the dislocation 
microstructure. 

In this work, a parametric model for generating synthetic training data for segmentation
of dislocations is developed. Even though domain scientists might dismiss synthetic 
training images sometimes as too artificial, our findings show that they can 
result in superior performance, particularly regarding the generalizing of the Deep 
Learning models with respect to different microstructures and imaging conditions.
Additionally, we propose an enhanced deep learning method optimized for segmenting
overlapping or intersecting dislocation lines. Upon testing this framework on 
four distinct real datasets, we find that our synthetic training data are 
able to yield high-quality results also on real images--even more so if fine-tune 
on a few real images was done. 

Our approach demonstrates the potential of synthetic data in overcoming the 
limitations of manual annotation in TEM, paving the way for more efficient and 
accurate analysis of dislocation microstructures. Last but not least, segmenting 
such thin, curvilinear structures is a task that is ubiquitous in many fields, 
which makes our method a potential candidate for other applications as well.

\tiny
\keyFont{ \section{Keywords:} deep learning, synthetic training data, segmentation, 
data mining, transmission electron microscopy, dislocation, crystal defect} 
 \normalsize
\end{abstract}

\section{Introduction}                                  \label{sec:introduction}
Crystalline defects play an important role for the performance and reliability
of many devices and components and are of big interest in materials 
science and materials physics. For example, the motion and interaction of 
dislocations, i.e., linear defects in the crystal lattice of metals or 
semiconductors, is directly responsible for plastic deformation and thereby has 
a strong influence on the resulting mechanical properties. Understanding the 
dynamic behavior of dislocations is therefore of great importance. One way
of achieving this is to observe them while they move and interact.
In-situ \TEM\ is a microscopy method that allows to do so and is the method
of choice for this work. 

\Ac{TEM} is the most often used  method to directly visualize dislocations. It is
based on the interaction of an electron beam with the crystalline specimen in
form of a thin foil. Further more, in-situ \TEM\ studies allow to
simultaneously perform mechanical testing and to capture the motion of such
defects as well as their interaction with obstacles, e.g., other dislocations,
second phase particles or grain boundaries \cite{Legros14}. 
\enquote{Quantitative in-situ \TEM} frequently refers to the ability 
of capturing both the evolution of a microstructure through electronic imaging 
and at the same time to estimate (externally) the associated stress response  as
already demonstrated in a number of studies
\cite{kiener2011source, yu2015situ,kacher2019impact,lee2020situ}. This
could -- in principle -- also help to understand the \emph{structure-property
	relationship} by extracting data from \TEM\ images (see, e.g., the work and 
discussion in \cite{Legros14}). Lee et al. \cite{lee2020dislocation} observed
the evolution of dislocation plasticity and additionally calculated the shear
stress acting on dislocations from the estimation of their line curvature in a
high entropy alloy. 
In all these studies, the stress is only roughly estimated because the local
line geometry, i.e., in particular the local curvature, could not be obtained.
Steinberger et al. \cite{steinberger2023data} manually extracted the position
of dislocations and used these as input for finite element simulations to study
the local stress state. Utt et al. \cite{utt2020jerky} tracked the positions of
dislocations over time and observed that they show jerky motion and move by
continuous pinning and depinning. While all authors could extract new and
interesting information from their investigations, only very few \TEM\ images from a
sequence could be analyzed as this had to be done manually. Therefore, the
statistical scatter from these \enquote{singular data points} is rather large
and can make it difficult to capture the entire dynamic process of dislocation 
motion appropriately. 

A prerequisite for analyzing the
dynamic behavior of the observed defect microstructure is the ability to extract
all dislocations as mathematical lines. These can be, e.g., represented by
polynomial approximations such that even local geometrical properties can be
represented as demonstrated in \cite{Sandfeld2015, Zhang2022}.
Digitization of dislocations is almost always done manually. This 
\enquote{hand labelling} was done in past without specific software, and only 
recently also dedicated tools such as \texttt{labelme} \cite{russell2008labelme}
were used. There, dislocations can be 
annotated as a line by selecting points on the dislocations.  In 
\cite{Zhang2022} this was used to extract the information of dislocations from 
300 individual frames where each frame has up to 20 dislocations. This is a 
challenging and tedious task, difficult to reproduce, and the result 
additionally may depend on the experience of the person who performs the
labelling of the images. The last aspect is a particular challenge as stress 
calculations can be very sensitive with respect 
to the local radius of curvature of a point on the line. With detectors and 
cameras becoming increasingly faster, there is an urgent need to automatize the
analysis of the huge amount of image data generated by \ac{TEM} in general and
during in-situ \TEM\ experiments in particular.  

\Acf{DL} based methods can be very powerful tools for performing  pixel wise
classification to segment objects of interest. This can significantly help to
automate the analysis of microscopic images 
\cite{minaee2021image,sasaki2022nanoscale}. In general, state of the art \DL\
architectures such as the U-Net \cite{ronneberger2015u} have been found to be
very successful for image segmentation with applications ranging from the 
classical field of  computer vision 
\cite{demir2018deepglobe, he2018learning, li2018deep} to medical imaging  
\cite{fu2020deep,litjens2019state,ward2018deep}. Such models also have been
applied to several problems in the field of material science, e.g., to segment
nano-particles  \cite{mill2021synthetic} and even for identifying
precipitates, voids and simple dislocation networks \cite{roberts2019deep}.
Furthermore, Shen et al.  \cite{shen2021multi} used \DL-based segmentation to
identify small defect loops.
While these methods are clearly very useful for analysis of images they suffer
from the problem that they require large amount of training data and/or are
not accurate enough to extract dislocations as mathematical lines. 
Usually in this scientific domain, training data is not available and needs to
be manually created to obtain the ground truth for a supervised
\ac{ML} task. 

An additional challenge specific to binary segmentation of dislocations
from TEM images is the variance of the observed ``dislocation phenomena''
along with varying imaging conditions: creating a representative amount of
training data that covers all possible configurations requires many experiments.
This is very expensive and time consuming; performing experiments that cover
rarely observed situations might even be impossible in all generality. A lack
of a diverse dataset often leads to over-fitting of the \ac{ML} model
where the already learned images are very well predicted but prediction for
micrographs from new experiments are surprisingly bad: thus, the generalization
of the trained model to new data/situations is strongly limited. These
difficulties are even amplified by the fact that there are no publicly
accessible data sets of \TEM\ dislocation microstructures available that could
be used to augment the training data set or that could be used in a transfer
learning approach. 

However, training on small datasets can be possible to some extent: Sasaki et
al. \cite{sasaki2022nanoscale} tried to circumvent the problem of 
\enquote{never-enough-data} and used the first 100 frames of
a \TEM\ video for training and the next 70 images for testing. Unfortunately, 
no quantitative evaluation of the performance was given. In another study
Roberts et al.  \cite{roberts2019deep} obtained two images of  $2048\times 2048$ pixels 
in size, divided the whole image into five parts and used three of them for
training, one for validation and one for testing. Together with basic augmentation
operations this resulted in a total training data set of 48 images. The
\ac{IoU} performance for dislocations was stated as 44\%. Nonetheless, care
needs to be taken when the test data used is very similar to the training data
as in such cases the model can again be prone to over fitting.   

The lack of high-quality \emph{and} high-quantity training data can be overcome 
by generating artificial images of dislocation microstructures -- an approach 
which so far has been only rarely used in the field of materials science.
There are several techniques available to generate synthetic data ranging from 
domain randomization \cite{tremblay2018training} where non realistic objects 
are added to force the \ac{ML} model to learn important features, to 
\ac{ML} guided methods as, e.g., \acp{GAN} 
\cite{thambawita2021singan} where a \ac{ML} model learns to generate 
synthetic data with features similar to real data. Chun et al. 
\cite{chun2020deep} used a \ac{GAN} to generate synthetic heterogeneous 
energetic material microstructures. 3D rendering software as, e.g., Blender 
\cite{blender2018blender}  can also be used to generate synthetic data. This 
is used in \cite{mill2021synthetic,cid2021deep} to create very realistically
looking images. 
Another possibility in the field of  material science is to use simulation 
methods to create \emph{synthetic images}. For this there are -- in 
principle -- many methods and models readily available, covering all phenomena
from the electronic scale up to macroscopic engineering scales. Such
simulations can be used to generate synthetic data as shown by Hajilounezhad
et al. \cite{hajilounezhad2021predicting} 
where a physics-based scanning electron microscopy (SEM) simulation tool is 
used to get artificial images of a carbon nanotube forest along with the 
calculated properties of the structures used as ground truth. 
Trampert et al. \cite{trampert2021deep} used Voronoi tessellations to generate
synthetic polycrystalline microstructures: even though the statistical
properties of the grain size distributions are only roughly comparable to those
reported for real grain distributions, the degree of visual similarity and the
type of contained features (e.g. triple points) turned out to be sufficient for
an effective training process. 
This demonstrates that synthetic data can be a very useful way of obtaining
training data for those cases where no suitable, \enquote{real} microscopy 
training data is available.  However, up to date, there exists no systematic 
analysis of how to create synthetic training data for TEM images of dislocation 
microstructures and how to 
infer which features need to be included and which are superfluous. \\

This paper is organized as follows: The parametric model to generate synthetic training data is presented in section \ref{sec:parametric_model}. The machine learning model and the loss function used in this work to predict a single dislocation in a single mask is described in section \ref{sec:ML_method}. The results are presented as well as discussed in section \ref{sec:Results} . The conclusions of the study is presented in section \ref{sec:conclusions}.

\section{Synthetic Data generation model} \label{sec:parametric_model}
In the context of microscopy, \emph{synthetic data} refers to images that mimic 
the \emph{relevant} details, i.e., geometrical and imaging features usually found 
in real microscopy images. Typically, other and less relevant details are then 
neglected, as is the case in any model of the reality. Finding out which aspects 
of \TEM\ images need to be considered and, based on that, how to generate 
effective synthetic training data is one of the goals of this work.  

\begin{figure}[htbp]
	\centering
	\includegraphics[width=\textwidth]{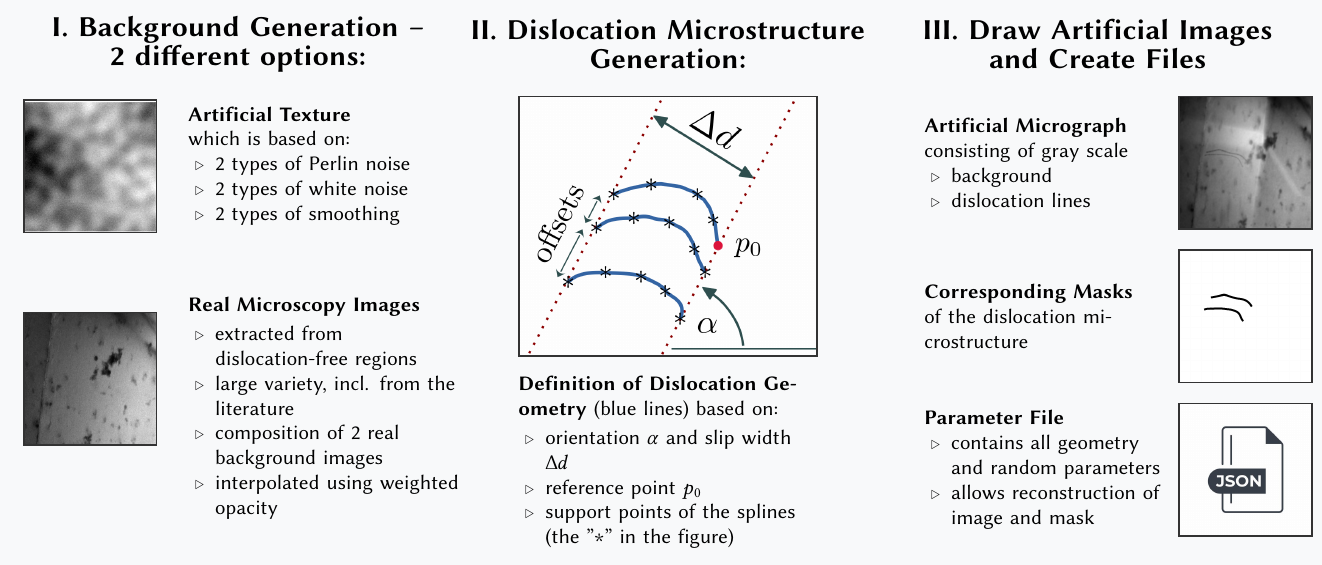}
	\caption{%
		Synthetic images and masks are generated in three steps. 
		First, a background image is created from artificial \enquote{noise
			generators}, or by using background images from real microscopy data
		Then, the dislocation geometry is determined and subsequently \enquote{drawn} on top of the background providing the synthetic
		image,  mask as well as the parameter file.  
	}
	\label{fig:three steps for synthetic images}  
\end{figure}

A parametric model for creating artificial \TEM\ images must satisfy two 
requirements: (i) create images and masks that are sufficiently similar to real 
images so that once a ML model is trained on the synthetic training data it can also provide high quality results on real images, and (ii) to use as few parameters as possible for describing the details of the image. The first
requirement is obvious while the second is important for easily adapting the
parametric model to other \TEM\ experiments or materials and thereby to enhance
generalization in situations \emph{beyond} the single experiment.
%
Creating a synthetic image of a dislocation microstructure consists of three 
concrete steps:
\begin{enumerate}
	\item[1.] 
	Generating a background for the synthetic image: 
	Two fundamentally different approaches are used in this work. One is
	completely artificial and superimposes different types of noise and
	smoothing operations. The second approach uses background patches from real
	microscopy images.
	
	\item[2.] 
	Generating the geometry of the artificial dislocation microstructure:
	there, the position and geometrical shape of the dislocations 
	and possibly other elements are determined.
	
	\item[3.] 
	``Drawing'': the dislocations on the background along with writing 
	the image and mask as a PNG file, and recording all parameters in a JSON 
	file for full reproducibility.
\end{enumerate} 
An overview of these steps together with a list of the most important 
parameters is shown in \cref{fig:three steps for synthetic images}. In what
follows, all steps and the involved parameters will be explained. 

\subsection{A Purely Synthetic Approach to Background Generation}  \label{sec:perlin_background}
Noise, randomness and a well-chosen general variance of different features in
the training dataset are key ingredients for a good training process; they reduce 
the risk of overfitting and are also helpful for generalization of the model to 
new images. Therefore, the generation of suitable background textures was an 
important task. For this purpose, we analyzed a number of real microscopy images
and found, that very often a number of different gray value gradients (e.g., 
owing to the microscopy imaging conditions) as well as random fluctuations in the 
brightness occurs. E.g., these may be due
to random \enquote{dirt} particles on the surface of the specimen or also due
to the presence of other defects that are not or only barely visible due to the
diffraction condition.

\begin{figure}[htbp]
	\centering
	\includegraphics[width=\textwidth]{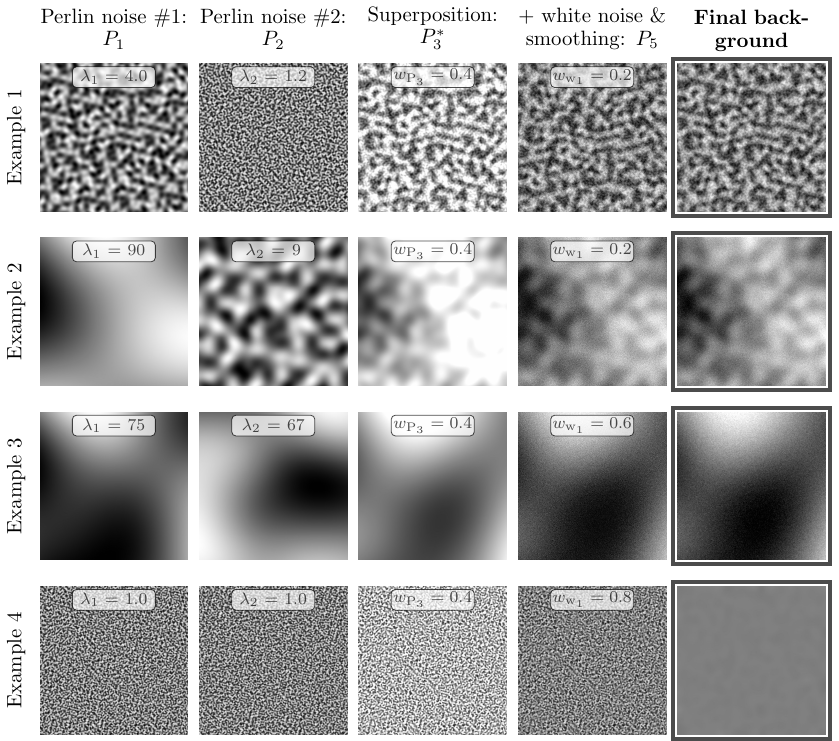}
	\caption{%
		The rows show four examples for synthetically created background
		images. The most important steps of the background creation pipeline 
		together with the used parameters are shown from left to right. 
	}
	\label{fig:background}
\end{figure}

We use two-dimensional Perlin noise which has been widely used to create 
texture for computer games 
\cite{wijgerse2007generating,inoue2021initialization,tatarinov2008Perlin}
because the structure is natural looking and can be designed such that it does 
not repeat itself within the image. The algorithm of Perlin noise uses random
gradients on regular grid points of the two-dimensional domain along with an
interpolations to generate random noise of a particular main wave length. 
For this work, a Python implementation from \cite{perlin_github} was used.
Perlin noise does not exactly look like the backgrounds of real \TEM\ images as
can be seen in the examples in \cref{fig:background}, and hence may not exhibit
the same statistical properties as the background of real images, e.g., in
terms of brightness distribution or the spectrum of wave lengths. However, it
offers an effective way of introducing a non-trivial type of randomness (as
opposed to, e.g., white noise). 
As a default, we use a superposition of Perlin noise with two different
wave lengths where the larger wave length is motivated by the mild gradient 
of gray value changes that encompasses the whole \TEM\ image, e.g., due to the
 imaging conditions. The smaller wave length represents all other 
fluctuations. Additionally, we applied a sequence white noise followed by a
Gaussian filter two times with different parameters.

We assume that the image has $M$ rows and $N$ columns. Two Perlin noise
distributions (or rather: arrays), $P_1(i,j)$ and $P_2(i,j)$ with $i=[1,M]$ and
$j=[1,N]$ are generated. $P_1$ has the dominant wavelength of ${\lambda}_{1}$,
$P_2$ has the wavelength ${\lambda}_{2}$. These distributions  are then
superimposed by weighting $P_2$ with a factor ${w}_{\rm P}$, resulting in
\begin{eqnarray}
	&P_3 =& P_1(\lambda_1) + {w}_{\rm P} \times P_2(\lambda_2)\\
	\Rightarrow\quad &P_3^* =& \frac{P_3 - \min(P_3)}{\max(P_3) - \min(P_3)} 
	\in[0,1] \quad \textrm{for all $i$ and $j$}\;.
\end{eqnarray}
where  $P^*_3(i,j)$ denotes the scaled and shifted  $P_3$. Next, white noise 
$X_1(i,j)$ is added by sampling each pixel value of the image array from a 
uniform random distribution with values in between $1$ and $-1$. The 
white noise is then weighted with 
${w}_{\rm w_1}$ and superimposed with the previous data to obtain 
\begin{eqnarray}
	\hphantom{\Rightarrow\quad}&P_4=&P^*_3 + {w}_{\rm w_1} \times X_1\;,
\end{eqnarray}
The resultant is (usually: only slightly) smoothed using a convolution with a
discrete Gaussian filter kernel $G_1$ that has standard deviation ${s}_{1}$:
\begin{eqnarray}
	\hphantom{\Rightarrow\quad}&P_5 =& G_1(P_4;s_1)\;
\end{eqnarray}
The steps $P_4$ and $P_5$ give slightly \enquote{smeared out} random
fluctuations which might, e.g., stem from the noise of the microscope electron
optics or the camera noise. 
We then add additional white noise $X_2$ weighted by the standard deviation
of the values of $P_4$, i.e., ${w}_{\rm w_2}=\sigma_{\rm P_4}$,  and apply one 
more Gaussian filter $G_2$ with standard deviation ${s}_{2}$ to adjust the 
width of 
these spikes, 
\begin{eqnarray}
	\hphantom{\Rightarrow\quad}&P_6 =& P_5 + {w}_{\rm w_2}\times X_2\\
	\hphantom{\Rightarrow\quad}&P_7 =& G_2(P_6; s_2)\;.
\end{eqnarray}
Due to the scaling of this final white noise, $P_7$ might also contain negative
values. Together with a \enquote{clipping} of the data to the range of $[0,1]$
this gives the final image $P$.
The smoothing of the last noise contribution typically changes the image only 
slightly and therefore gives sharper, high frequency fluctuations, in particular
together with the clipping. The combination of the last two noise types along 
with smoothing is also a way to mimic artifacts from the usually lossy image 
compression often used in proprietary microscopy software.

Our studies shows that the two types of Perlin noise together with some
high frequency oscillations are the most effective contributions for a 
high quality training dataset and even only one white noise contribution would
often suffice. To increase the variance of the dataset we nonetheless kept both 
white 
noise contributions together with the subsequent smoothing operations.
In \Cref{fig:background} we show some examples for intermediate steps and the 
final background together with the respective parameters. The images created
have a resolution of $512\times 512$ pixel.
%
%
%

In the first two 
examples only the wave lengths of the Perlin noise contributions are varied.
There, it can be observed that textures with very different wave length 
and characteristics can be reproduced. Some of them resemble backgrounds 
of real microscopy images, but not all of them do. Below, we also will study 
the importance of the degree of realism.

All parameters as well as the seeds for the random number generators were 
recorded and stored in a JSON file so that the synthetic background could be 
fully reconstructed.

\subsection{Background Generation Using Real Images} 
\label{sec:background:real}
The second approach to generating background images for our study is based on utilizing real images with the objective of producing highly realistic background textures. To achieve this, we have gathered a total of 170 \TEM\ images featuring dislocation microstructures, specifically focusing on larger regions devoid of dislocations. These "background-only" areas were then carefully cropped and extracted by hand. The images were obtained from a diverse range of sources, including an extensive body of pre-existing literature. It is important to note that we have not used any backgrounds from the four real datasets employed in this study. To enhance the diversity of the data, two background images were chosen at random and superimposed with a random opacity value. Although the resulting superimposed image may appear somewhat unusual or "incorrect" to an expert observer, the presence of additional background features compared to a typical real background has proven beneficial for the generalization of ML model when applied to real data. In order to ensure the reproducibility of this background generation process, we have meticulously documented all parameters of the individual images, as well as the opacity parameter used during the superimposition process. This will enable other researchers to replicate our methodology and build upon our findings in their own scientific investigations.

\subsection{Generation of Artificial Dislocation Microstructure}
Dislocation structures as observed in real \TEM\ images can strongly vary in 
terms of shapes, relative sizes with respect to the image, the number of active
slip systems or the orientations of dislocations (some of such examples are
shown in \cref{fig:4real_datasets}). 
Throughout this work, we mainly consider dislocations in so-called pileup configurations 
(this denotes nearby dislocations that strongly interact with each other and
often move roughly together in groups). There, the special case of a single 
dislocation is understood as a pileup having only one dislocation. 

When dislocations move they create so-called \emph{slip traces} corresponding 
to the interception of the glide plane of these dislocations with the free surfaces 
of the thin foil. In an in-situ \TEM\ experiment they appear as weakly contrasted, 
dark or light straight lines that should be ignored by the \DL\ model. The slip 
traces are characterized by the angle
of inclination, $\alpha$, and by the slip width $\Delta d$  as shown in 
\cref{fig:three steps for synthetic images} and are used for 
determining the dislocations' positions.
A dislocation starts at one of the slip traces and ends at the
other slip trace line as shown in \cref{fig:three steps for synthetic images}II. The start point of the first dislocation in a group (pileup)
of dislocations is given by the parameter $p_0$. Here, the dislocation line is 
mathematically represented as cubic spline, and the geometrical shape of
the line is governed by a number of support points. These support points for a dislocation can be obtained 
either from labeling dislocations in real \TEM\ images using e.g., the software
tool \texttt{labelme} \cite{russell2008labelme} or by prescribing some
reasonably looking coordinates (e.g., by randomly choosing an average curvature
and then adding random variations to each of the support points).
Once the points for the first dislocation have been obtained we can move to the next dislocation of the pileup. The next dislocation in such a group can then be obtained by
shifting the first dislocations' support points along the slip traces by an 
offset value, followed by adding some 
randomness to the final line shape or by prescribing new set of support points as before. For any subsequent dislocation, these steps are repeated, possibly by using different offset values.

A single real microscopy image may contain several groups of dislocations
that might move into different directions. This can be easily realized by simply
repeating all above steps, starting with determining the slip traces. 
Again, we record all used parameters as well as the seed values for the random
number generator and store them in a JSON file.

\begin{figure}[h]
	\centering	\includegraphics[width=13cm]{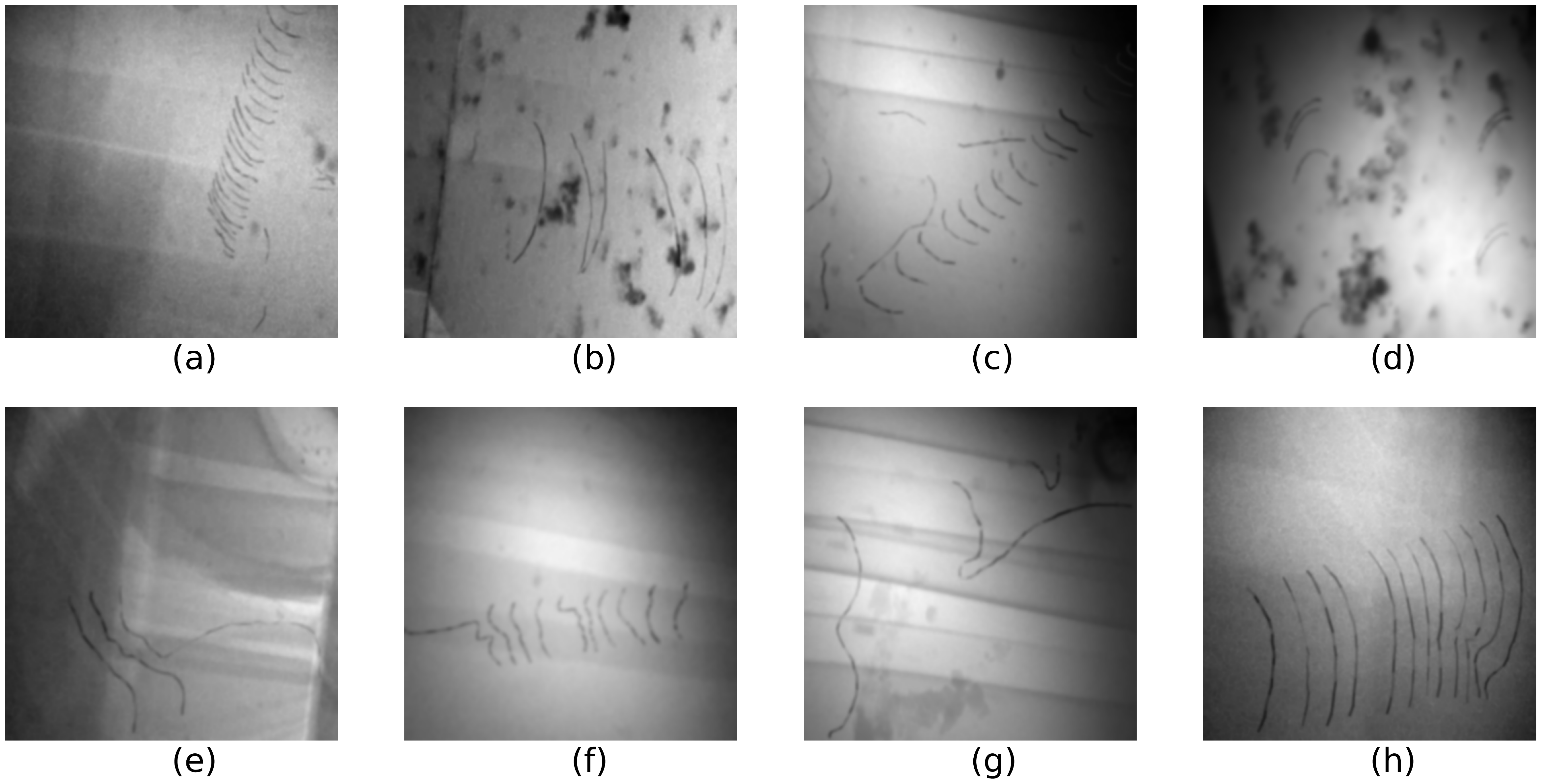}
	\caption{Some of the wide range of synthetic images that can be generated by different values of the parameters.}
	\label{fig:sample_synthetic_images}
\end{figure}

\subsection{Drawing images and masks}
As a final step, we use the Python package \emph{Matplotlib} \cite{hunter2007} 
to draw the dislocations on top of the respective background. Each dislocation
line has a gray value and line thickness that is randomly chosen from a range
of reasonable values, e.g., we ensure that the dislocation appears slightly 
darker than the background, as is also the case in real images \cref{fig:4real_datasets}).

While drawing the gray dislocation lines on the background we also draw the 
dislocation lines as solid black lines on an empty canvas for creating the
ground truths for the segmentation task. Again, we save all required parameters
as JSON file for easy reproducibility.
The synthetic images do not contain the same physical basis which govern real
dislocation microstructures and may even violate physical laws. However, the 
synthetic images only serve as training data for the machine learning task and 
therefore the most important aspect is that the data contains enough parameters
and features to be generalized to other or even more complex dislocation
structures.
The synthetic image generator can be used to create  synthetic images with a 
wide range  of dislocation microstructures. Some of the synthetic images are 
shown in \cref{fig:sample_synthetic_images}. 

\section{Machine learning method} \label{sec:ML_method}
Our objective is to segment dislocations which can be post-processed to represent each segmented dislocation as a mathematical spline. Converting dislocations into splines enables us to perform quantitative calculations, such as computing the velocity, position, and curvature, as demonstrated in \citep{Zhang2022} for a selected, hand-annotated frames.  Training a machine learning model for segmentation tasks can be challenging, in particular when predicting masks that maintain pixel spacing between closely situated dislocations, as illustrated in \cref{fig:4real_datasets} where real images (a) and (c) have very close dislocations with barely any pixel spacing between them. If there is no pixel spacing between two dislocations, it becomes difficult to assign unique dislocation identifiers to the pixels and hence makes it difficult to represent them as dislocation splines. This issue is further complicated in images with intricate dislocation microstructures, where dislocations overlap and intersect one another. 


\begin{figure}[htbp]
	\centering
	\includegraphics[width=\textwidth]{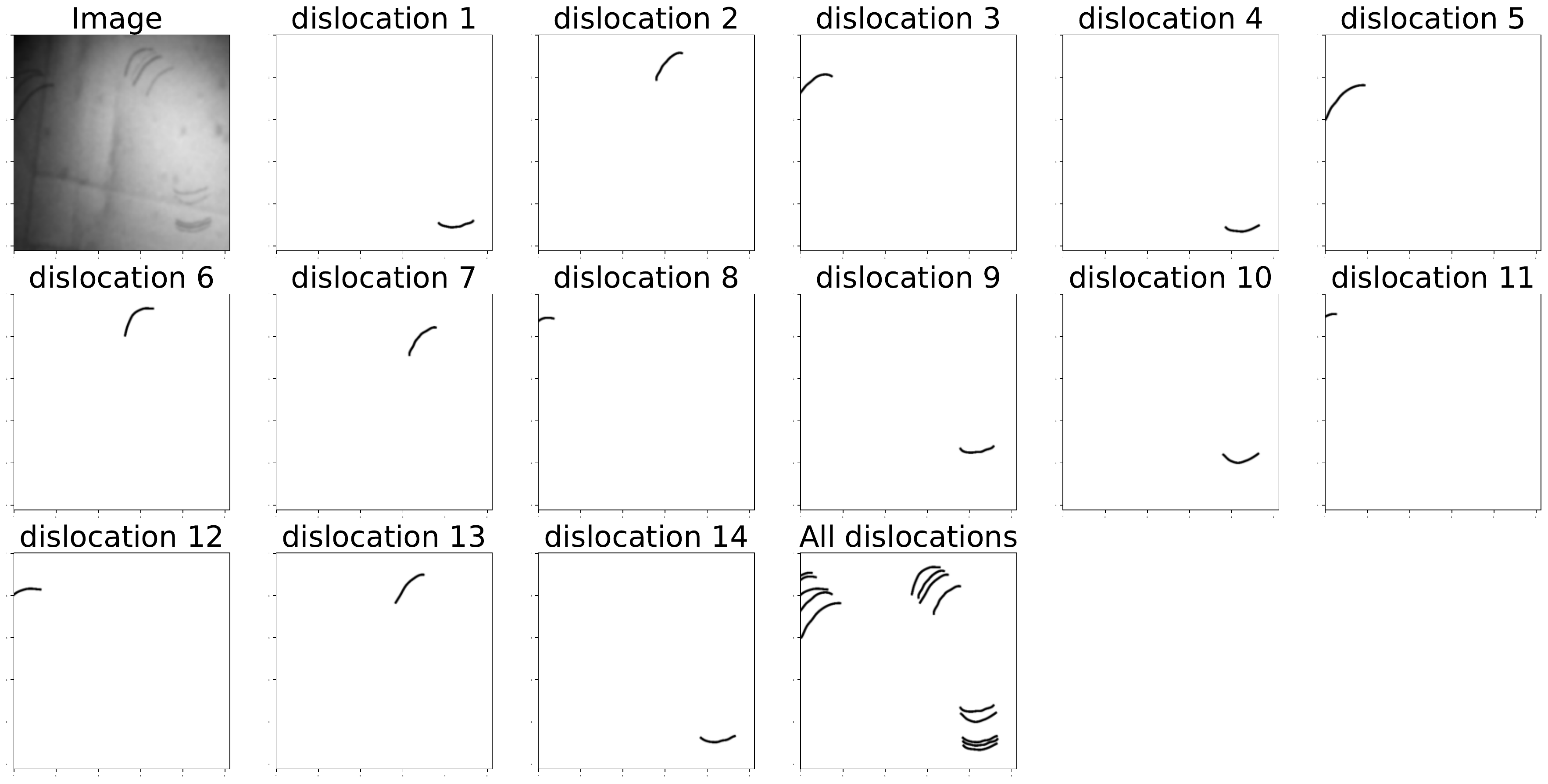}
	\caption{%
		A synthetic image of a dislocation microstructure which has 3 dislocation pileups along with the 14 dislocations present in the image which are shown in individual masks. If the masks of all the 14 dislocations are combined we can obtain all the dislocations in a single mask as shown in the all dislocations image.   
	}
	\label{fig:sample-training-data}
\end{figure}

We present a methodology that aims to solve this challenge by  predicting dislocations as separate instances, with each instance denoting a single dislocation in an individual mask, thereby ensuring that all pixels within a mask correspond to the same dislocation. We have used a U-Net++ model with a ResNet50 backbone, which is implemented using the PyTorch library \cite{Iakubovskii:2019}. Although more advanced architectures exists, the primary focus of this research is on the use of synthetic training data. Thus, benchmarking alternative network architectures is not part of this
work; only a small study was conducted that showed that the used architecture is a good 
compromise between accuracy and computational efficiency. 

The input to our model comprises images of $512\times512$ pixels in size, and the output is a set of masks with dimensions $N\times512\times512$ pixels, where $N$ is a parameter for the maximum estimated number of dislocations within an image. In the examined datasets, we have estimated $N$ to be 20 which will allow the model to predict a maximum of 20 masks for each of the 20 dislocations. 

We now take a look at a first scenario of segmentation of dislocations. An example of a synthetic image along with the corresponding masks, is depicted in \cref{fig:sample-training-data}. This particular image represents a complex scenario of three dislocation pile-ups, which together contain a total of 14 dislocations. The role of the machine learning model in this context is to analyze this image and predict a total of 14 masks, each containing only a single dislocation out of the 20 total masks considered as shown in the figure. Ideally, the six extra masks that we have considered for the sake of accommodating any potential extra dislocations should be predicted as empty by the model. However, for the sake of simplicity in this work, we also accommodate the possibility of these extra masks containing dislocations, with the understanding that any duplicate dislocations can be conveniently filtered out in a subsequent post-processing step. This is likely the case for images which might have only as few dislocations (1 to 8). In such a case, multiple masks may predict the same dislocation but they can be easily filtered during a post processing step. Furthermore, our investigation of the synthetic image reveals that several of the dislocations are extremely closely spaced, in some cases almost to the point of having a sub-pixel distance between them. This high-density distribution of dislocations presents a particularly challenging segmentation task which can be tackled easily with the proposed method. 

In order to train the model to predict single dislocations within each mask, we propose a novel loss function, which is an adaptation of the widely-used Dice loss. The calculation of this loss for an image is presented in \cref{pseudocode}. For an image containing $M$ dislocations, the model generates $N$ instances of dislocation predictions. Since the instances in the prediction $PM$ and ground truth $GT$ are not ordered, we must first determine the correspondence between predicted and actual dislocations. We achieve this by computing the Dice loss between the instances of the ground truth and predicted dislocation masks. We start with the first ground truth mask which will be used to calculate the dice loss with all the predicted masks. This gives us $N$ values of the dice loss. We find the predicted mask with the minimum dice loss which would corresponds to the first ground truth mask. 

Once we have found the corresponding predicted mask for a ground truth,  we can calculate metric to evaluate performance of the model. In this work we have introduced a new metric to evaluate the model's performance, specifically tailored to dislocation image data. The evaluation method diverges from traditional techniques such as Dice or Intersection over Union (IOU), and instead processes dislocations in the masks directly as mathematical splines. The procedure involves binarizing the predicted mask, post-processing the mask, applying Lee skeletonization \cite{au2008skeleton} which converts the length objects to one pixel thickness. These points can also be treated as support points for the spline to represent the dislocation. Finally we calculate the relative error between the length of the predicted dislocation and the length of the ground truth dislocation. Although this metric follows as very strict criteria and may yield lower scores due to minor prediction errors i.e., in cases where a part of the dislocation is not predicted, it provides a more accurate reflection of the model's performance, particularly in terms of the quality of the predicted masks to postprocess them to represent each dislocation as a spline. A value of one for the metric would represent a case where the mask is predicted not only very accurately but when the predicted dislocation is represented as spline the spline has the same length as the dislocation in corresponding true mask. 

Once the loss and metric value for the first true mask has been calculated, we remove them from list of all predicted and true masks and repeat the process for the remaining dislocations. This process gives us  $M$ Dice losses corresponding to each dislocation which is averaged to get the overall loss and metric for the image.

\begin{algorithm}
	\begin{itemize}
	\item GT: Ground Truth Masks (1,2,...,M)
	\item PM: Predicted Masks (1,2,...,N)
	\item CDL: Calculate Dice Loss
	\item CMI: Corresponding Mask Index
	\item L: Average Loss of an image
	\end{itemize}
	\caption{Loss Calculation for an Image}
	\label{pseudocode}
	\begin{algorithmic}[1]
	\Require $GT$, $PM$, $M$, $N$
	\Function{ComputeLoss}{$GT$, $PM$, $M$, $N$}
	\State $L \gets 0$ \Comment{Initialize the average loss}
	\For{$i \gets 1$ to $M$} \Comment{Iterate through ground truth masks}
	\State $DL_{min} \gets \infty$ \Comment{Initialize minimum Dice Loss}
	\For{$j \gets 1$ to $N$} \Comment{Iterate through predicted masks}
	\State $DL \gets \Call{CDL}{GT[i], PM[j]}$ \Comment{Compute Dice Loss}
	\If{$DL < DL_{min}$} \Comment{Check if the current Dice Loss is smaller}
	\State $DL_{min} \gets DL$
	\State $CMI \gets j$ \Comment{Update corresponding mask index}
	\EndIf
	\EndFor
	\State $L \gets L + DL_{min}$ \Comment{Update the average loss}
	\State Remove $PM[CMI]$ \Comment{Remove matched predicted mask}
	\EndFor
	\State $Loss \gets \frac{L}{M}$ \Comment{Calculate the final loss}
	
	\State \Return $Loss$
	\EndFunction
	\end{algorithmic}
\end{algorithm}


In this work we perform training on synthetic datasets and  generate 4000 training images and use 1000 images for testing. During the training of our models we try to achieve optimal performance on the synthetic datasets. To ensure the best model performance, we saved the model that achieved the highest score based on our physics-based metric evaluated on the synthetic test data for each of the synthetic datasets. This is a well known method to prevent overfitting and improve generalization of unseen data. We use a number of image transformation methods during the training process such as applying Gaussian noise, changing brightness, contrast and image equalization to provide a wide range of synthetic images with different texture properties to improve generalization to real images.
\section{Real data}              \label{sec:realdata}
\begin{figure}[h]
	\centering
	\includegraphics[width=13cm]{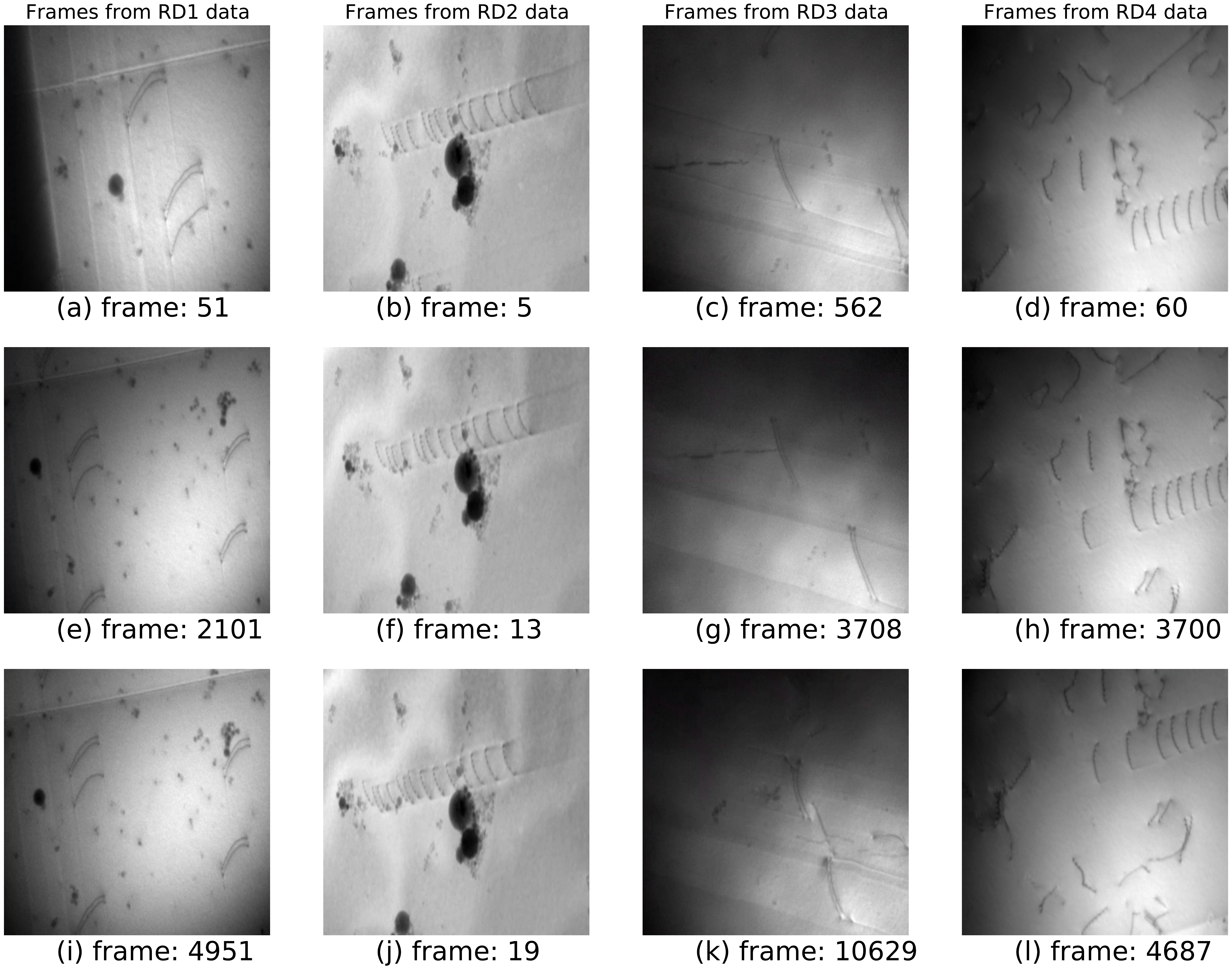}
	\caption{3 different frames from 4 real datasets named as RD1,RD2,RD3 and RD4 obtained from different experiments. }
	\label{fig:4real_datasets}  
\end{figure}

We use data from four in-situ TEM experiments, named as RD1, RD2, RD3, and RD4 to test the ML model trained on synthetic datasets.  We extracted the frames from the experimental videos using their respective frame rates and each experiment produced thousands of frames, with the primary differences between consecutive frames being the position and shape of the dislocations. These variations in dislocation microstructure can be observed in \cref{fig:4real_datasets}, where frames from the same video are shown. We can see that snapshots within the same dataset are relatively similar and only details of the dislocation structure, the camera position and the lightning conditions changes. Furthermore the four datasets
themselves are fairly different. Among the various dynamic sequences, we tried to work on typical low density dislocation configurations that are those where the cleanest observations are made. Those are also the ones that are usually analyzed "manually" by expert microscopists because Burgers vectors, slip planes and local curvature can be easily extracted from the images. One of the challenges here was to discriminate dislocations that can be closer to each other while others are more easily resolved. This can happen when a pile-up is stopped on a strong obstacle and in some alloys where short range order may induce such pairing \cite{saada2004pile}.

Each dataset comes with different challenges. One major challenge is to resolve the nearest dislocations which can be seen clearly in data RD1 and RD3 where two dislocation can be very close to each other. An other challenge which is most common is incorrect prediction of a slip trace line as dislocation. For some data, i.e., RD3 the slip trace line can be very distinct and the model may predict a part of a slip trace line as dislocation. For RD2 and RD4 it might be more challenging to predict so many dislocations as single instances in the masks. Furthermore, there is an additional challenge in RD4 which contains many dislocations of different sizes and shapes. 

\section{Results and Discussion} \label{sec:Results}

\subsection{Synthetic training data}

The challenge of determining suitable parameter values for synthetic data generation models 
results from the need to ensure that a \ac{ML} model trained on synthetic images can also yield satisfactory results when applied to real images. The following factors contribute to the complexity of this task:

\begin{enumerate}
    \item Data distribution disparities: Synthetic images may not accurately capture the distribution, variations, or intricacies present in real-world images. As a result, an ML model optimized on synthetic images may struggle to generalize effectively to real images.

    \item Incomplete representation of real-world conditions: Synthetic image generation models may not encompass all potential scenarios or conditions found in real images, such as varying lighting, occlusions, or textures. This limitation can lead to a ML model that is ill-equipped to handle dislocation segmentation tasks of real images.

    \item Challenges in parameter selection: Identifying the optimal combination of parameters for a synthetic data generation model is a complex and resource-intensive process. The difficulty increases if the ML model has to be adapted to account for discrepancies between synthetic and real images.
\end{enumerate}

Selecting appropriate parameter values for synthetic data is difficult, 
primarily due to the fact that synthetic images are always more or less "different" from real images. Nonetheless the synthetic images provides a unique alternative due to the cost and time involved in generating a representative and diverse real TEM images of dislocations. 
In many instances, the actual data remains inaccessible during the training phase, making it impossible to utilize it for modeling synthetic data. To address this, we generate two distinct synthetic dislocation microstructures, M1 and M2, based on whether the real data is utilized to obtain parameter values  to generate synthetic data or not.

\begin{figure}
	\centering
	\includegraphics[width=13cm]{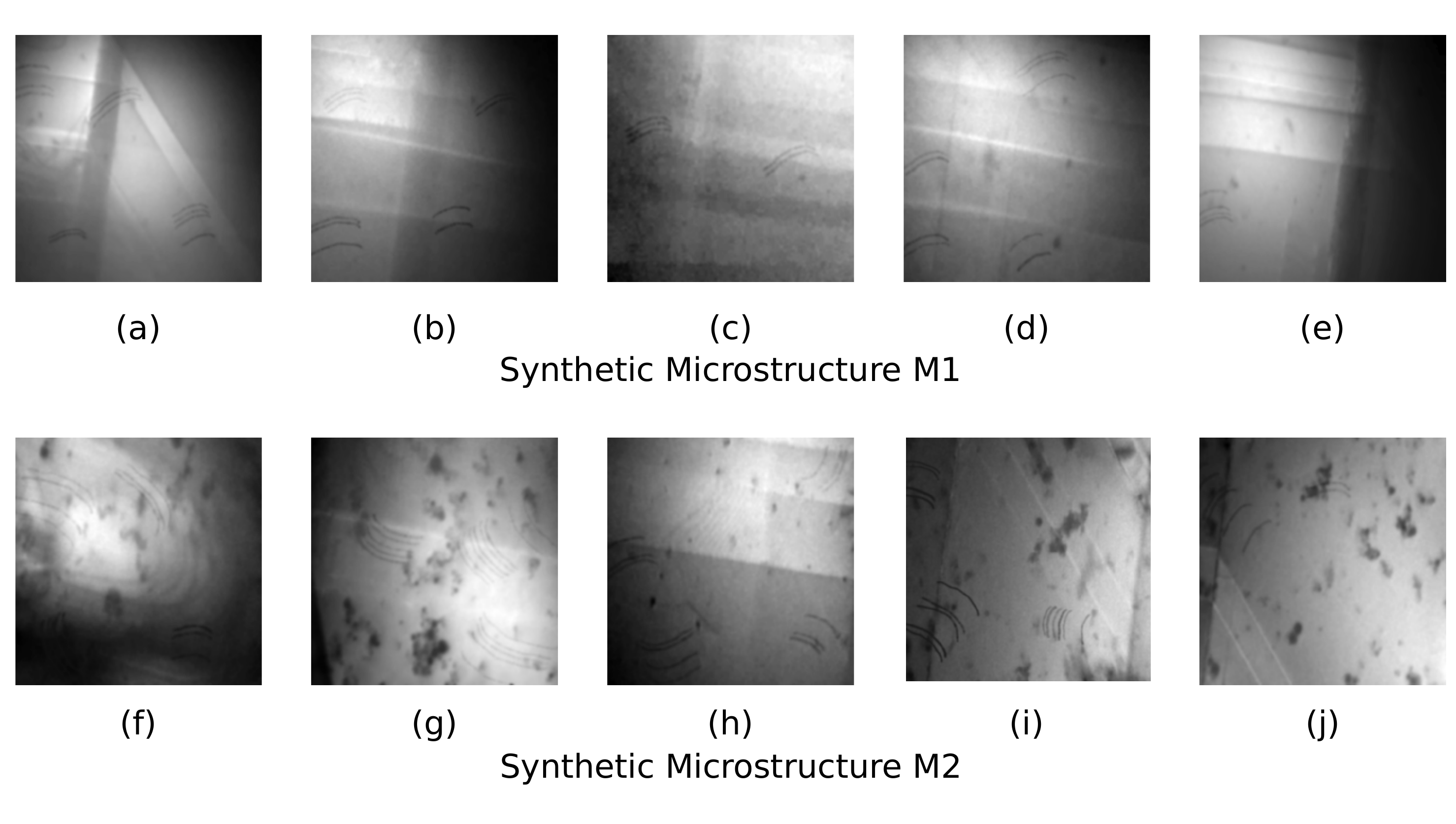}
	\caption{Sample synthetic image for microstructure M1 and M2}
	\label{fig:sample_images_RD1}
\end{figure}

\begin{figure}
	\centering
	\includegraphics[width=\textwidth]{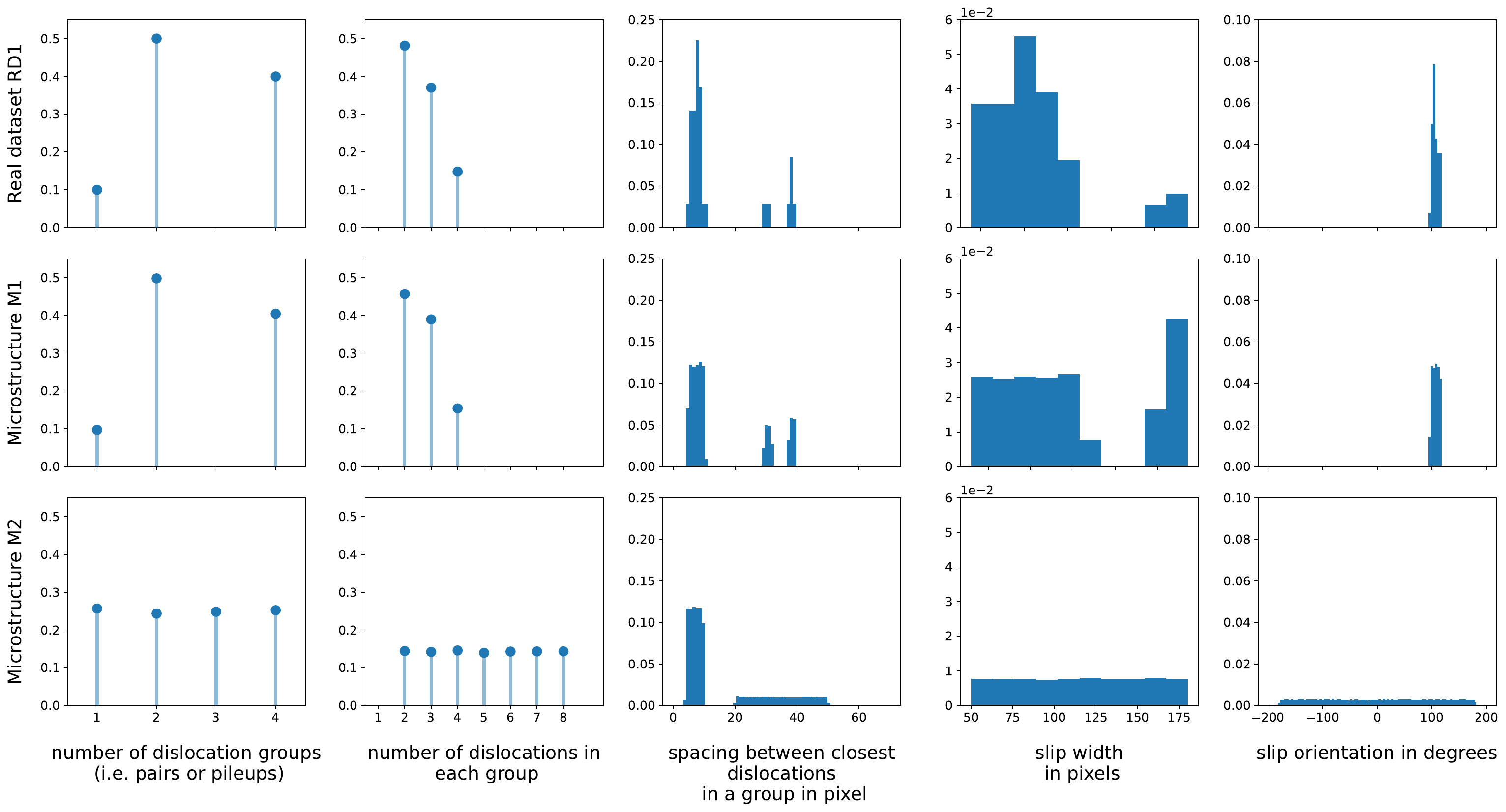}
	\caption{Feature distributions of Synthetic microstructure M1, M2 along with the Real data RD1. The microstructure M1 is modeled based on Real data RD1. }
	\label{fig:microstructure_feature}
\end{figure}

We use real data, RD1, to simulate the microstructure M1 of synthetic data. 15 hand-labeled images from the RD1 dataset are used to determine the distribution of various microstructural features, such as the number of pileups, the number of dislocations within a pileup, slip width, slip direction, and the spacing between the two nearest dislocations in a pileup. The synthetic data generation model can also create a "synthetic twin" of an actual image by replicating its dislocation microstructure. However, to ensure variation in the training data, we have not duplicated the exact microstructure of the real data but rather sampled the parameters according to the distribution of microstructural features. The probability distribution of these features for the real data and synthetic microstructure M1 is shown in \cref{fig:microstructure_feature}. A selection of synthetic images with microstructure M1, is presented in \cref{fig:sample_images_RD1}. 
The synthetic images are very similar to the real images in terms of microstructure. On a closer look one can easily distinguish real and synthetic images which would make it difficult to generalize the ML model trained only on synthetic images. Here, the image transformation methods described in \cref{sec:ML_method} were very useful to generalize the trained  ML models to real images. 

In some situations the real data is unknown before training our model. In such a situation it is difficult to estimate suitable values of parameters to generate synthetic images. This motivated us to design another synthetic microstructure, M2 which is much more general, and we have used a "reasonable" range of values of parameters which are not motivated by any specific, real microstructure. 
The probability distribution of the microstructure features are shown in \cref{fig:microstructure_feature} where
we see that the distribution is much more general and uniform. There is only one parameter, "support points for dislocation splines" which is motivated from real data, though. Support points for dislocation splines in both the microstructures are obtained from dislocations in the real data RD1. 
Some of the images based on microstructure M2 are also shown in \cref{fig:sample_images_RD1}. 
On comparison we can see that these synthetic images have quite different microstructure than those in the real dataset RD1. There may be multiple pileups in the microstructure M2 each having different slip directions  with as many as 8 dislocations. Furthermore we can also see that in an image itself we can find multiple pileups having different slip directions. The synthetic images also have pileups with different slip widths. Looking at the sample images for microstructure M2, we can see that the microstructure is much more general compared to dataset RD1 or microstructure M1 itself. 

We use the two background generation methods described in \cref{sec:perlin_background} and \cref{sec:background:real} to generate the background of the synthetic images with the 2 microstructures which gives us four synthetic datasets. The most suitable parameters for the  Perlin noise were experimentally obtained and are shown with their value ranges  in \cref{Perlin}. 

\subsection{Machine learning results on synthetic data}
\begin{figure}[htbp]
	\centering
	\includegraphics[width=0.9\textwidth]{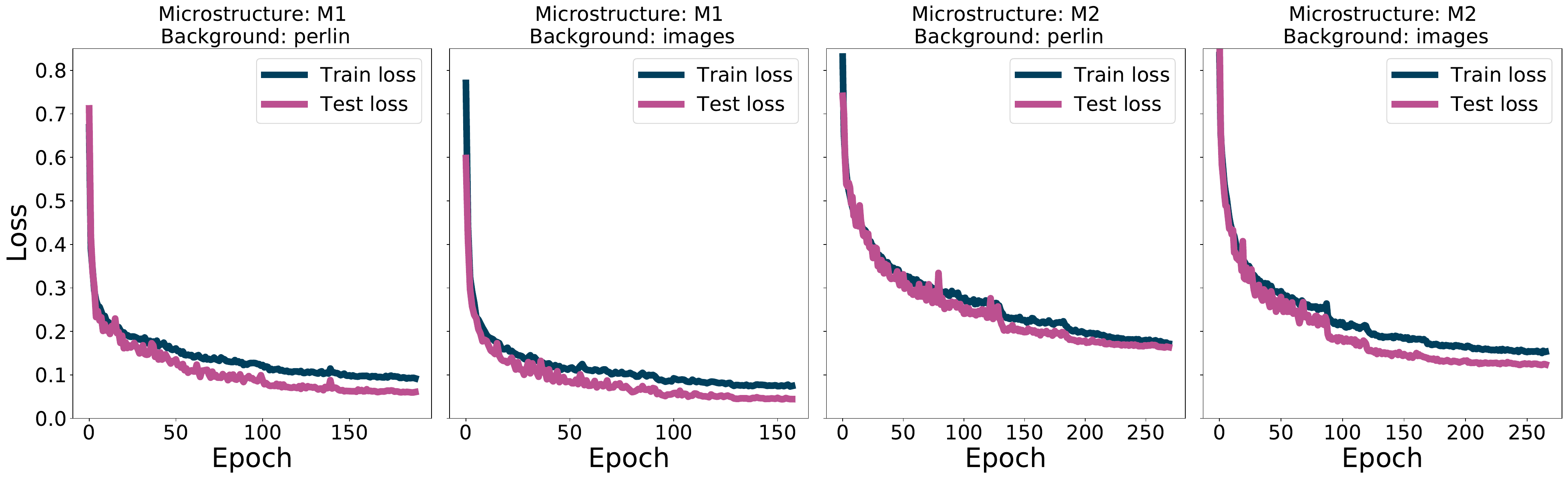}
	\caption{Training and test loss curves for training on the 4 synthetic datasets}
	\label{fig:loss_curves}
\end{figure}

 The training and test loss curve for the models trained on the four synthetic datasets are shown in \cref{fig:loss_curves}. Both the training and test losses decrease during the initial training phase. This is a typical behavior for well-trained models, as the optimization process should lead to a reduction in the loss function value over time. The loss curves reach a saturation level where the loss of the test curve is lower than that of the training curve. This observation  was consistent across all four synthetic datasets, further strengthening the claim of good generalization on the test synthetic dataset. The diverse nature of the training data allows the models to learn the underlying patterns and structure in the images, which in turn makes them more robust and adaptable to variations in the test images.

\begin{figure}[htbp]
	\centering
	\includegraphics[height=10cm]{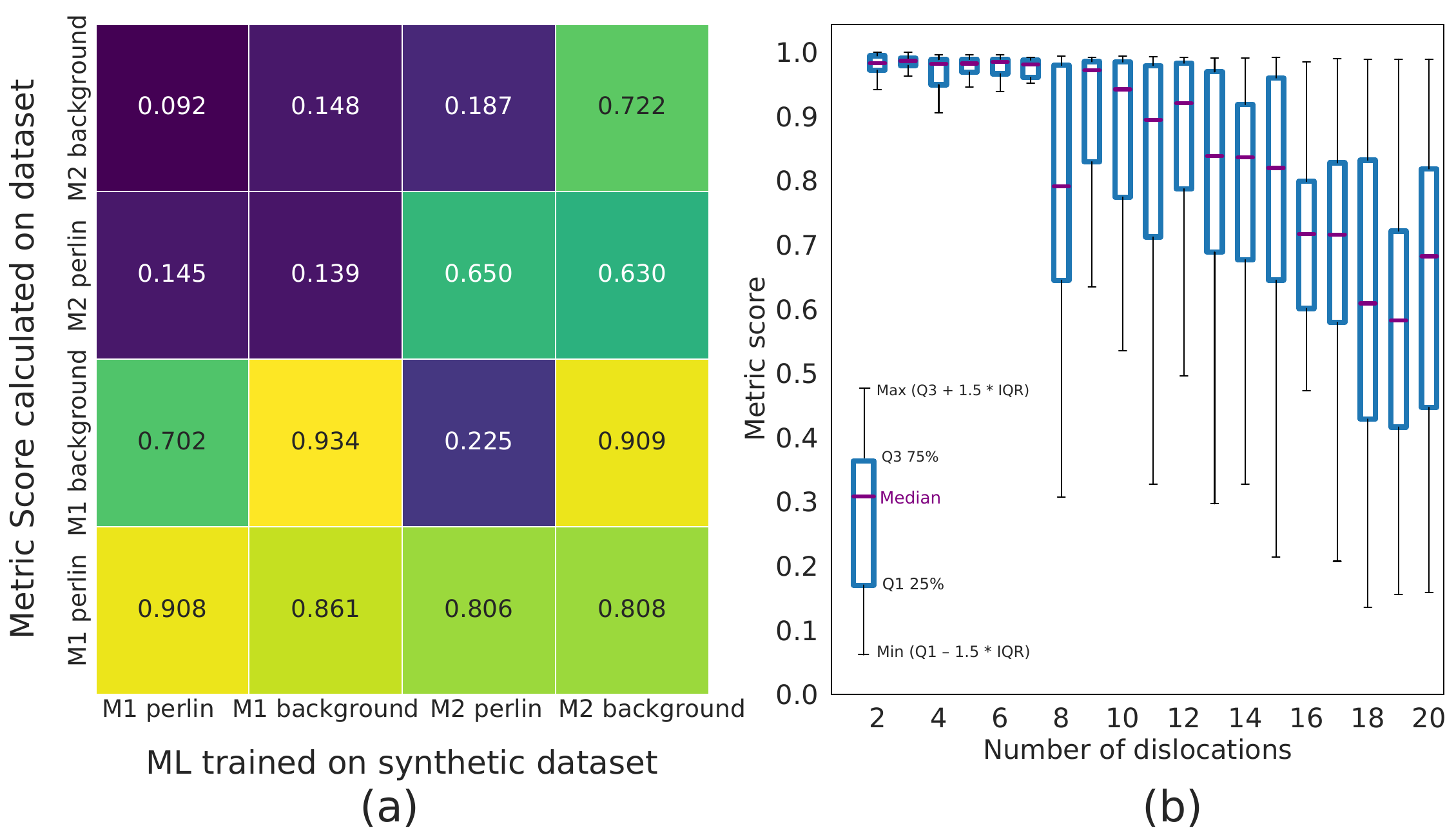}
	\caption{(a) Metric evaluation of the 4 trained models on 4 synthetic datasets. (b) Distribution of metric score of the ML model trained on dataset with microstructure M2 with realistic background with the number of dislocations present in the images.}
	\label{fig:4models4dataset}
\end{figure}

We also compared the performance of the ML model trained on one synthetic dataset against the other three datasets. We generated 2000 images for each dataset and then evaluated the metric of the four trained models on all four datasets. The results are shown in \cref{fig:4models4dataset} in the form of a evaluation matrix. On closely comparing the loss curves for the two microstructures, M1 and M2 we find that the ML model optimized easily on microstructure M1 where we were able to obtain loss on test dataset even lower than 0.1 as seen in \cref{fig:loss_curves}. The dataset with the general microstructure, M2, is difficult to optimize compared to the much simpler microstructure M1. We also observe this by comparing the diagonal values in \cref{fig:4models4dataset}a where the ML models trained on synthetic datasets with microstructure M1, give metric values as high as 0.9 on datasets with same microstructure but ML models trained on microstructure M2 could only give metric value of 0.722 on the dataset with M2 microstructure. 

We found that the models trained on synthetic data with Perlin noise do not generalize well to dataset where realistic background images were used, i.e., models trained on microstructure M2 with Perlin noise results in a very low score of less than 0.25 on synthetic data with background images. This is not the case with ML models trained on realistic background images such as dataset M2. We obtained metric values $>0.6$ on synthetic data with Perlin noise background which is comparable to the values obtained on real microscopy backgrounds. Thus, using realistic backgrounds is a better choice than using purely synthetic background. 

On comparing the results on the four datasets, we find that the ML model trained on the dataset with microstructure M2 and realistic background images generalizes well w.r.t. the other three datasets. The ML model not only gives good results with metric values as high as 0.8 on other microstructure but also on datasets with Perlin background.

We observed that for images with large numbers of dislocations it becomes difficult to obtain a mask that contains just a single dislocation. The distribution of the metric scores with the number of dislocations present in an synthetic image is shown in \cref{fig:4models4dataset}b. Obtaining metric values as high as 0.9 is possible for images with just a few dislocations (4-8), however, for images with larger numbers of dislocations ($>$12) the metric scores tends to get lower on average with increasing numbers of dislocations.
Nonetheless even for images with large number of dislocations i.e., 19 dislocations, the model could still obtain scores as high as 0.9 but the median is much lower. This is not a result of poor segmentation but rather stems  from the  difficulty in predicting only a single dislocation in a mask. As a rule of thumb, if  we find that multiple dislocations are predicted in a single mask then  the metric value becomes much lower. Nonetheless, we note that in all these cases, by combining predictions from all masks we can obtain a single mask that contains \emph{all} dislocations predicted in the image.

\subsection{Predictions on real TEM images}
While it is certainly reassuring that the performance on synthetic data is very good,
it is important to evaluate the performance of the trained ML models on real images. We take the best ML model, the one trained on synthetic images with realistic backgrounds with microstructure M2 and use it to make predictions on real images. In the following we discuss the results on each of the four real datasets. The model predicts a total of 20 masks for an image, some of which might not contain any dislocations at all.

\begin{figure}
	\centering
  	\includegraphics[width=\textwidth]{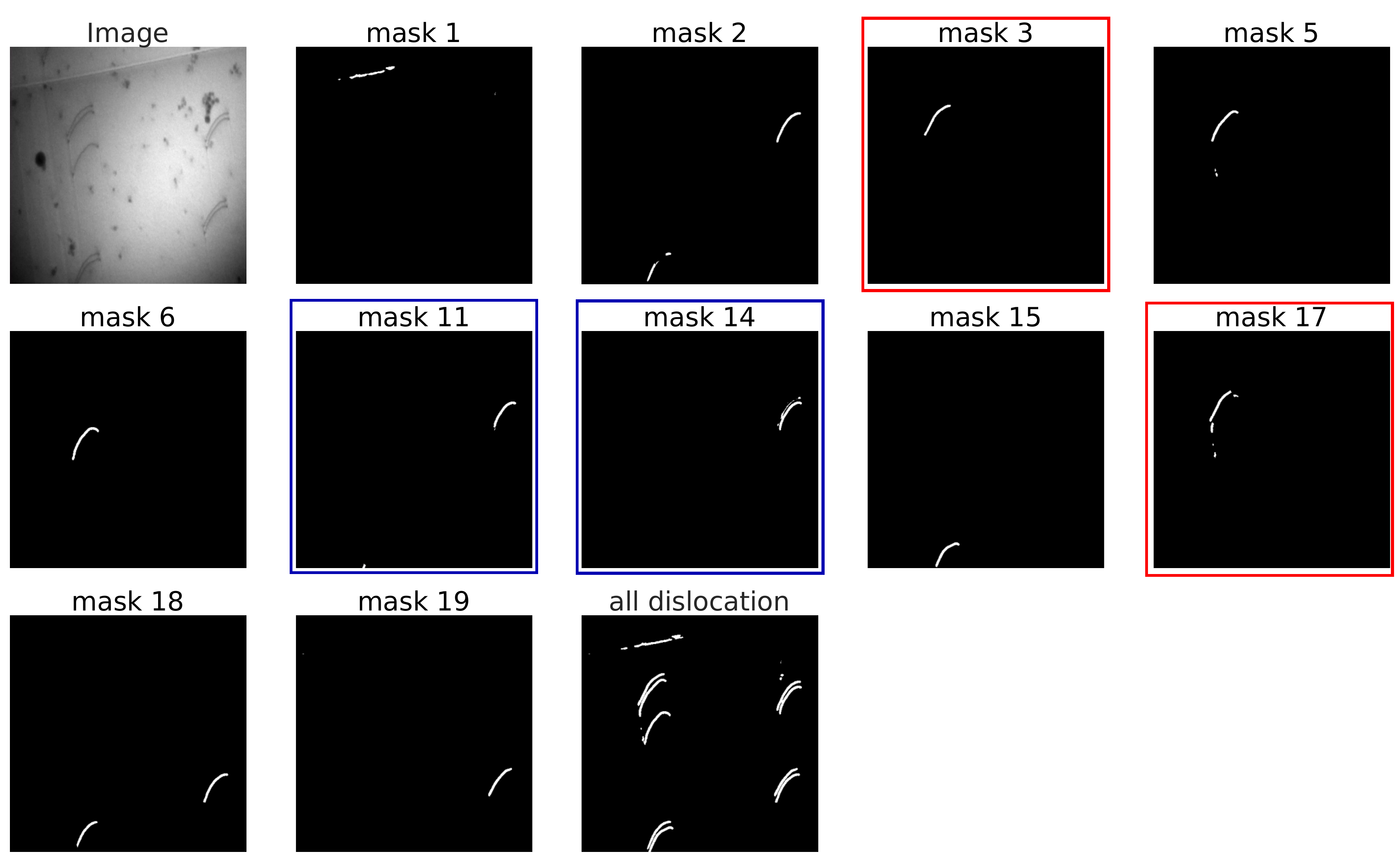}  
	\caption{Prediction on a real image from dataset RD1. Masks which predict the same dislocation 
	have titles shown in the same color (e.g., mask 11 and 14).}
	\label{fig:RD1}
\end{figure}

The model's predictions on a frame containing nine dislocations from the dataset RD1 are presented in Figure \cref{fig:RD1}.  For visualization purposes, empty predicted masks are omitted. For the sake of convenience we also exclude masks which contain only a very small number of ``dislocation pixels'' or some other, easy to detect irregularities. These are still visible in the combined output of all 20 masks which is shown in ``all dislocation''. Overall, the achieved accuracy is very good where the model was able to predict all the 9 dislocations present in the image. There are a few issues especially in mask 1, a line between two regions of different contrast (a slip trace or boundary to a twinned region) is incorrectly predicted as dislocation.  
Given that the image contains nine dislocations and the model can predict a maximum of 20, several masks may represent the same dislocation, such as masks 11 and 14, and 3 and 17 which has been highlighted with red and blue color. It is interesting to note that sometimes one of such multiple masks is much better than the other(s), e.g., the prediction in mask 17 also shows a small part of slip trace line (the vertically arranged pixels) while in mask 3 the prediction is much more accurate. To identify multiple masks representing the same dislocation, the Dice loss is calculated, and if the loss is found to be less than 0.5, the masks are considered to contain the same dislocation. This criteria is found to work well for all the multiple masks representing same dislocation. Further analysis reveals that mask 18 contains more than one dislocation. Nonetheless we can see that the proposed method worked quite well in predicting single instances of dislocations. This might have been more difficult for binary segmentation of dislocation where we might not have any pixel spacing between two very close dislocations. We can see that the results are not perfect in the sense that we have multiple masks predicting same dislocation and a mask containing more than one dislocation but these issues can be solved quite easily by post processing of these machine learning results.

\begin{figure}
	\centering
	\includegraphics[width=\textwidth]{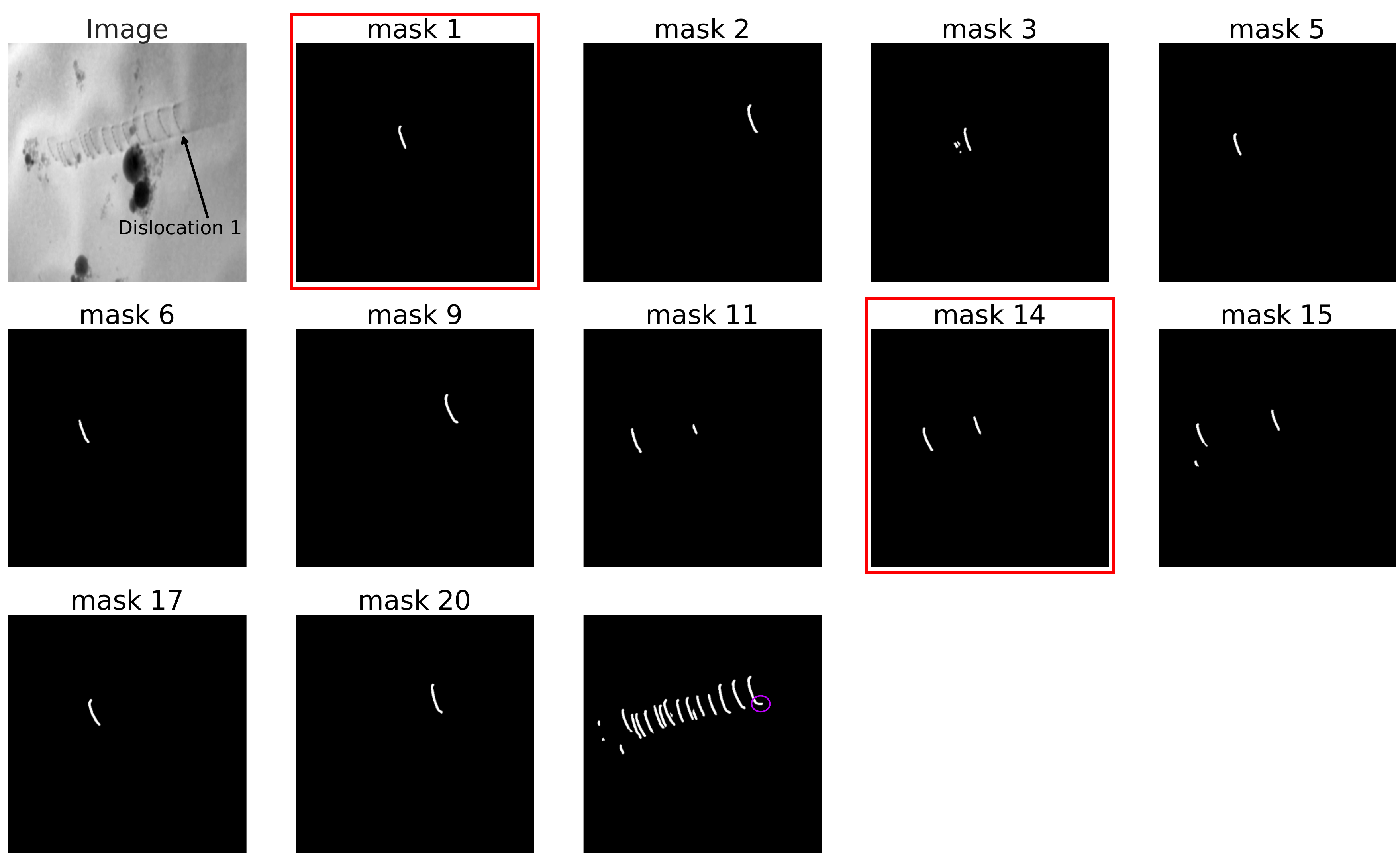}
	\caption{Prediction on a real image from dataset RD2. Dislocation named as id 1 is marked with arrow in the image. }
	\label{fig:RD2}
\end{figure}

The second example is based on the real microscopy dataset RD2 with  \cref{fig:RD2} showing a snapshot along with the predictions. The dislocation microstructure consists of a pileup of 14 dislocations. We find that for this real image, the model was able to predict eight dislocations in a single mask but masks 11, 14, and 15 contains more than one dislocation either completely or partially. Furthermore we can see in the combined mask that the model fails to predict one of the dislocation. On a closer look we see that the first dislocation from the right (highlighted as dislocation 1 in the figure) is predicted quite accurately in mask 2, however, the same dislocation in the combined mask has on the lower end an extra short, horizontal segment (the circled part). This artifact stems from a ``slip trace'' that was incorrectly also predicted as dislocation. We can see here that the image in this dataset is very different than RD1 but still the model was able to predict the dislocations quite well. Qualitatively analyzing the results we conclude that they still require some improvement.

\begin{figure}
	\centering
	\includegraphics[width=\textwidth]{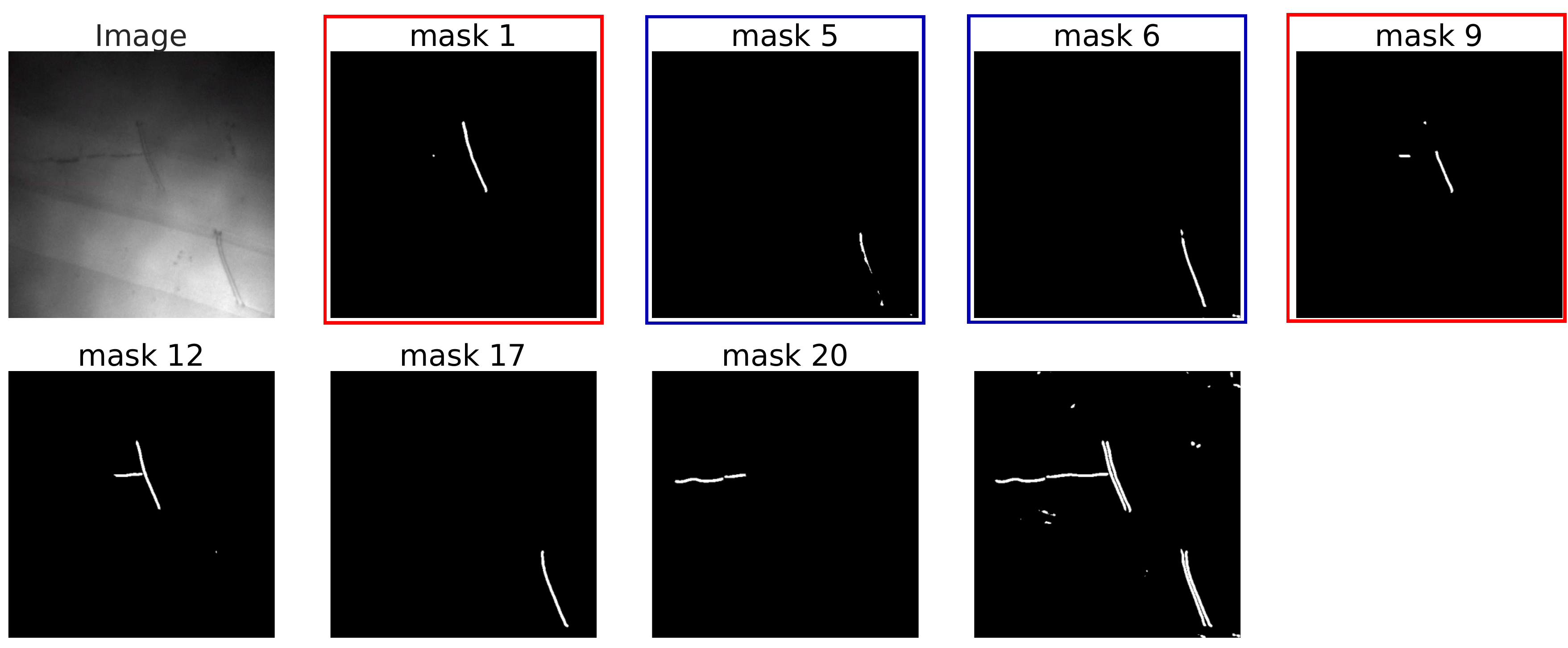}
	\caption{Predictions on a real image from dataset RD3.}
	\label{fig:RD3}
\end{figure}
Dataset RD3 focuses on a difficult to segment situation: two pairs of dislocations which move together and have a very small distance, cf. \cref{fig:RD3}. The model is able to predict all four dislocations very accurately. There is one single dislocation present which is ``stuck'' in the crystal in horizontal direction. This dislocation is also predicted but only through composition of two masks namely 12 and 20. 
We also find that since the image has less dislocation than the number of masks, there are multiple masks which predict the same dislocation. Mask 1 and 9 represent the same dislocation but on comparison we see that that dislocation in mask 1 is the accurate prediction. This is one of the shortcoming of the model: when multiple masks represent the same dislocation, it is difficult to automatically identify the mask that accurately represents the dislocation. This is a typical example of a real image where two dislocations are so close to each other that there is hardly any pixel spacing between them. Altogether, the model works quite well despite the fact that the distance between the dislocations is extremely small. With a small amount of post-processing one can easily extract the dislocation position and geometry.

\begin{figure}
	\centering
	\includegraphics[width=\textwidth]{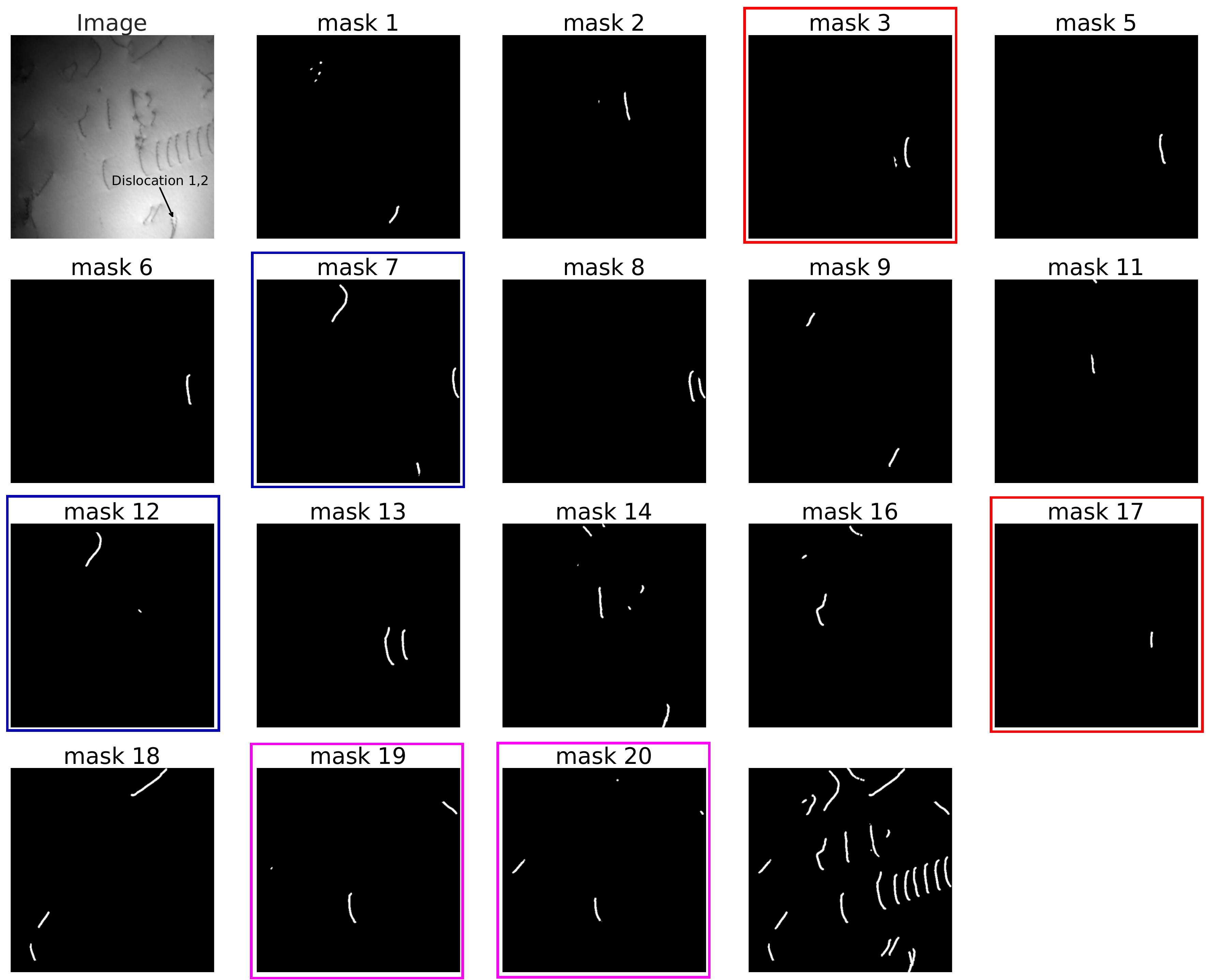}
	\caption{Predictions on a real image from dataset RD4. Two very close dislocation named as dislocation 1 and 2 are marked with arrow in image.}
	\label{fig:RD4}
\end{figure}
%
As a final example in dataset RD4  we now turn to a situations of a very complex dislocation microstructure that was not part of the synthetic training data. The dislocation microstructure in this image consists of a large number of pileup of dislocations in different slip directions and with different slip width along with a number of dislocations which are isolated and rather randomly distributed.  The image has dislocations of different shapes. During the synthetic data generation process we did not considered such diverse dislocation shapes might be a challenge.  
We observe that the model was still able to predict most of the dislocations present in the image. Taking a look at two very close dislocations which overlap (marked as id 1 in the image) we see that one of the dislocations was predicted accurately in mask 14 and the second dislocation was (partially) predicted in mask 7. The prediction in separate masks is beneficial for identification and post-processing of the dislocations.
As the image contains more than 20 dislocations there are several masks that contain more than one dislocation, which is a limitation of the approach of using a fixed number of masks but which can also easily be remedied.

The results of the ML model which is trained on the general synthetic dataset, on the four real datasets looks very promising. We used a general range of the parameters to generate the general synthetic data as described which is very useful since we do not need to extract or process any real dataset to obtain the values of parameters. The results can be generalized to more real datasets by using a more general range of parameters along with a large variety of dislocation shapes which we might find. Even though in our work we have obtained the shape of the dislocations from real dataset RD1, we find the the trained model still worked quite well in predicting dislocations of other shapes i.e., dislocations in RD4. This could still be further improved by generating the synthetic data using more diverse shapes of dislocations.  

\begin{figure}
	\centering
	\includegraphics[height=4cm]{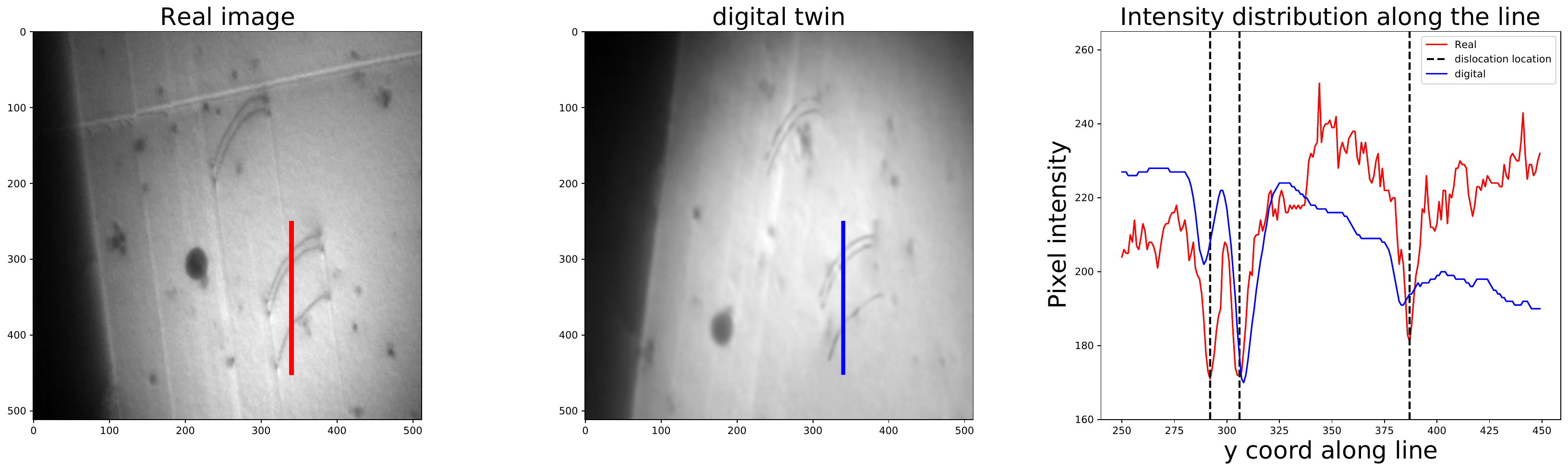}
	\caption{Comparison of Pixel intensity distribution around a dislocation between a real image and its digital twin.}
	\label{fig:histogram}
\end{figure}

To further investigate this difference in quality of the results on real images, we select an image from dataset RD1 and generate a ``digital twin'' by replicating its main microstructure features as far as possible with our synthetic training data generator, as illustrated in \cref{fig:histogram}. 
Although the backgrounds of the synthetic and real images differ, the dislocation geometries are consistent. We examined the pixel intensity distribution around dislocations by comparing the distributions along lines depicted in the images. Our findings indicate that the distribution for the digital twin is considerably smoother around the dislocations. This suggests that the method used to generate synthetic dislocations could still be improved, potentially leading to a more realistic pixel distribution around dislocation regions. Enhancing the quality of synthetic images could be beneficial for the generalization of the trained models on real images. By refining the process of drawing dislocations in synthetic images, we can more closely mimic the characteristics of real dislocation microstructures, making the synthetic images more representative of actual scenarios. The predictions of the ML model for both images are shown in \cref{app:digital twin} where we can see that the results on the synthetic image are much better compared to the predictions on real image.

\subsection{Application of the methodology}
\begin{figure}[h]
	\centering
	\includegraphics[width=\textwidth]{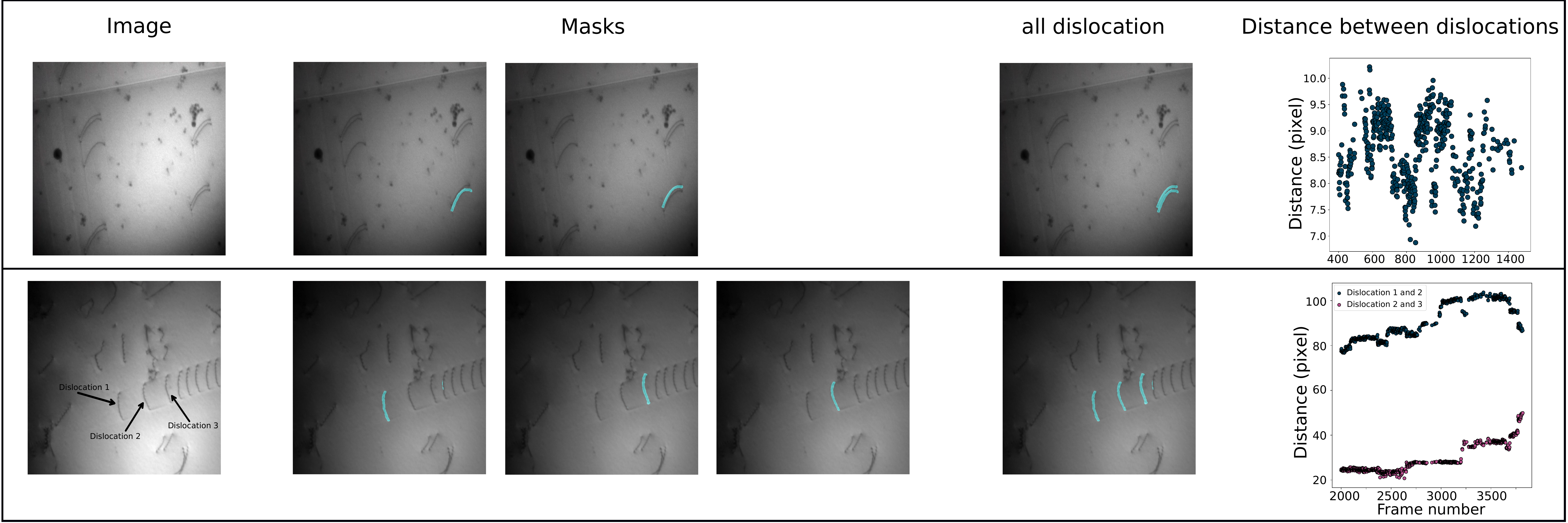}
	\caption{Extracting information about distance of dislocations for datasets RD1 and 4: shown are (from left to right) the real image, the predicted masks, the composition and the resulting plot of the distance vs frame number.}
	\label{fig:RD1-distance}
\end{figure}

In what follows we use a simplified approach to show how the predictions could be used to perform quantitative studies on the whole TEM image dataset. We use two real microscopy datasets, RD1 and RD4, to calculate the time evolution of distance between a pair of dislocations over several frames. 
A frame from each dataset is shown together with the respective predictions in \cref{fig:RD1-distance}.
RD1 consists of a number of dislocation pairs. We now calculate the distance between the two dislocations forming a pair and extract this information for all frames of the movie. All other dislocations are neglected. The dislocation microstructure for dataset RD4 contains more dislocations. Of interest are in particular the leading three dislocations of the pileup marked as Dislocation 1,2 and 3 in the figure.

To be able to extract individual dislocations and their geometry one can either perform image post-processing and try to mend some not-perfect masks. An alternative is to perform finetuning on just few hand-labelled data which can help us obtain nearly-perfect masks and can make extracting individual dislocations very accurate. Since our goal is to demonstrate the general usefulness of the whole methodology, we decided to take the latter approach where we take 10 hand labelled images from each dataset to find the ML model trained on the synthetic datasets.
%
%
For predictions on images from each of the two dataset, we can see that the fine tuned model was able to predict each dislocation in a separate mask much more accurately. This helps us tracking the dislocation through several frames. Each mask can then be binarized using a threshold of 0.5 which is followed by a Lee skeletonization to convert the pixel group to just one pixel thickness. This allows us to get points on the dislocations, and we interpolate all points to get equally spaced 50 points on each dislocation. The average distance between two closest dislocation can be calculated using all 50 points on the dislocations. The distribution of the distance as a function of the frame number is shown in the two scatter plots in \cref{fig:RD1-distance}. For RD4 we calculate the distance between the leading and the following dislocation along with the next dislocation. Such studies on the whole TEM data can help  domain specialist understand how dislocations move (smooth or "jerky"), about the strength of their interaction as well as their interaction with the material. Such studies can make important contributions towards understanding why materials behave the way they do..

\section{Conclusion} \label{sec:conclusions}

In this study we have presented a synthetic data generation model which can provide high quality training data for segmentation of dislocations. We have also presented a deep learning approach which successfully segments individual dislocations (where each individual dislocation should be located
in an individual mask). In our study we found that a general synthetic dataset which consists of a wide range of dislocation microstructures is useful concerning the generalization to completely new microstructures. The model is trained only on synthetic data and we were able to obtain high quality results on synthetic data. Real microscopy data was, as expected,  more difficult to predict but the results were still very convincing, given the complexity of some of the images. In particular, the individual segmentation of pairs of very nearby dislocations is an extremely challenging task for any ML model. The proposed method successfully addresses this issue by distinguishing between individual dislocations and providing more accurate predictions and identification of individual dislocations. The results highlight the model's ability to overcome the limitations of traditional binary segmentation approaches, making it a valuable tool for dislocation analysis as demonstrate by a small proof-of-concept example.

\section*{Conflict of Interest Statement}
The authors declare that the research was conducted in the absence of any 
commercial or financial relationships that could be construed as a potential 
conflict of interest.

\section*{Author Contributions}
Kishan Govind: Methodology, Software, Writing (original draft);
Daniela Oliveros: Investigation (TEM experiments);
Antonin Dlouhy: Resources (TEM samples);
Marc Legros: Conceptualization (TEM experiments), Writing (review $\&$ editing), Supervision;
Stefan Sandfeld: Conceptualization, Writing (original draft), Supervision, Funding Acquisition. 
All authors read and approved the final version of the manuscript.

\section*{Funding}
The authors KG, DO, ML, SS acknowledge financial support from the European Research Council through the ERC Grant Agreement No. 759419 MuDiLingo (“A Multiscale Dislocation 
Language for Data-Driven Materials Science”).

\section*{Data Availability Statement}
Most of the datasets along with Python code used in this study are available from zenodo \url{https://zenodo.org/record/...}. Datasets that can not by shared publicly are available upon reasonable request from the authors. 

\appendix
\section{Appendix}
\subsection{Used parameters for Perlin noise}
The  table in \cref{Perlin} shows the choice of parameters that  were used for creating synthetic backgrounds.
\begin{table}[h]
	\begin{center}
		\small
		\def\arraystretch{1.2} 
		\begin{tabular}{ 
				!{\extracolsep{1em}}
				>{\centering}p{1.25cm}
				>{\centering}p{2.8cm} 
				>{\centering\arraybackslash}p{3cm} 
			}
			\hline
			ID & variable name  & value range \\ 
			\cline{1-1}\cline{2-2}\cline{3-3}
			1 & ${O}_{1}$       & $10.0 \,\ldots\, 100.0$\\
			2 & ${O}_{2}$       & $10.0 \,\ldots\, 100.0$ \\ 
			3 & ${w}_\textrm{perlin}$  & $0.2 \,\ldots\, 0.4$ \\
			4 & ${w}_\textrm{white}$   & $0.5 \,\ldots\, 0.7$ \\ 
			5 & ${s}_{1}$       & 1. \\ 
			6 & ${r}_\textrm{noise}$   & $0.2 \,\ldots\, 0.4$  \\
			7 & ${s}_{2}$       & $0.6 \,\ldots\, 0.8$ \\ \hline
		\end{tabular}
	\end{center} 
	\caption{%
		Parameters and their values used for generating background images with 
		Perlin noise.
	}
	\label{Perlin}
\end{table}

\subsection{Prediction for a real image and its synthetic replication}\label{app:digital twin}
\begin{figure}[H]
	\centering
	\begin{minipage}[b]{0.48\textwidth}
		\includegraphics[width=\textwidth]{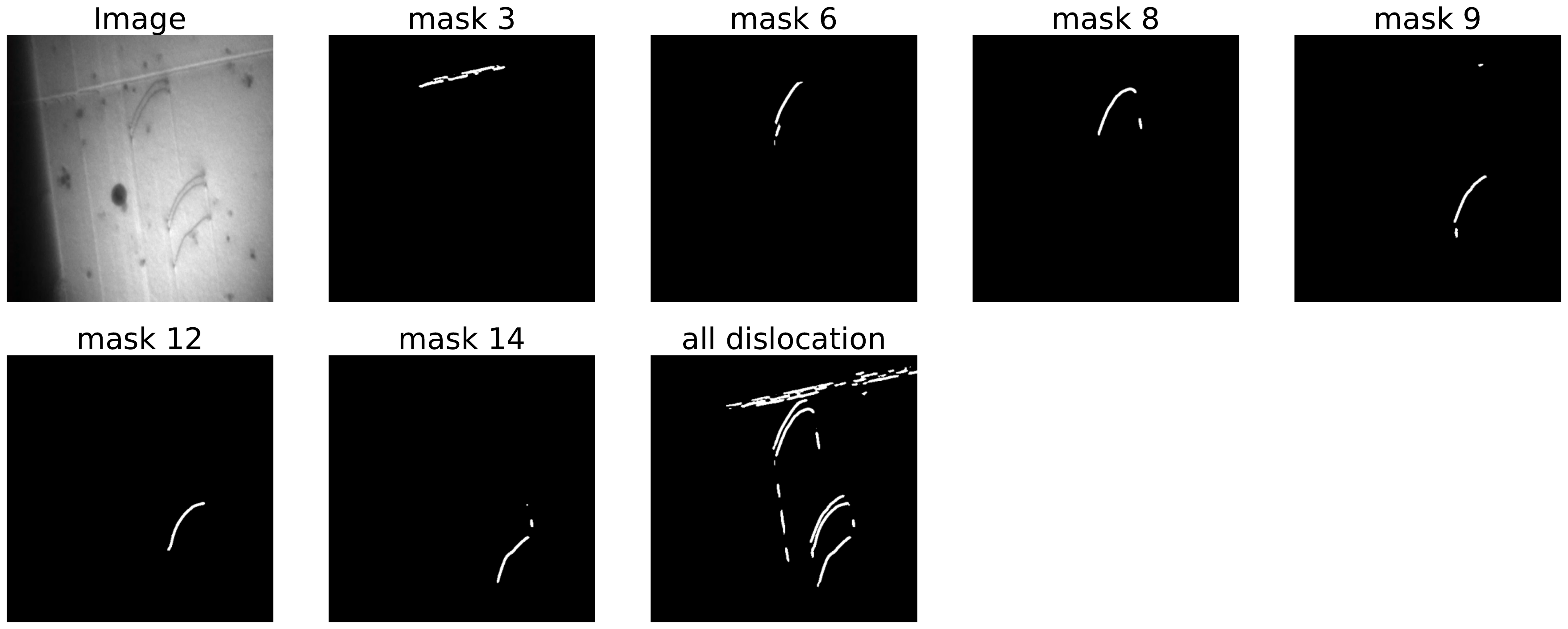}
    	\captionsetup{labelformat=empty}
		\caption{Prediction for real image from RD1 dataset}
	\end{minipage}
	\hfill
	\begin{minipage}[b]{0.48\textwidth}
		\includegraphics[width=\textwidth]{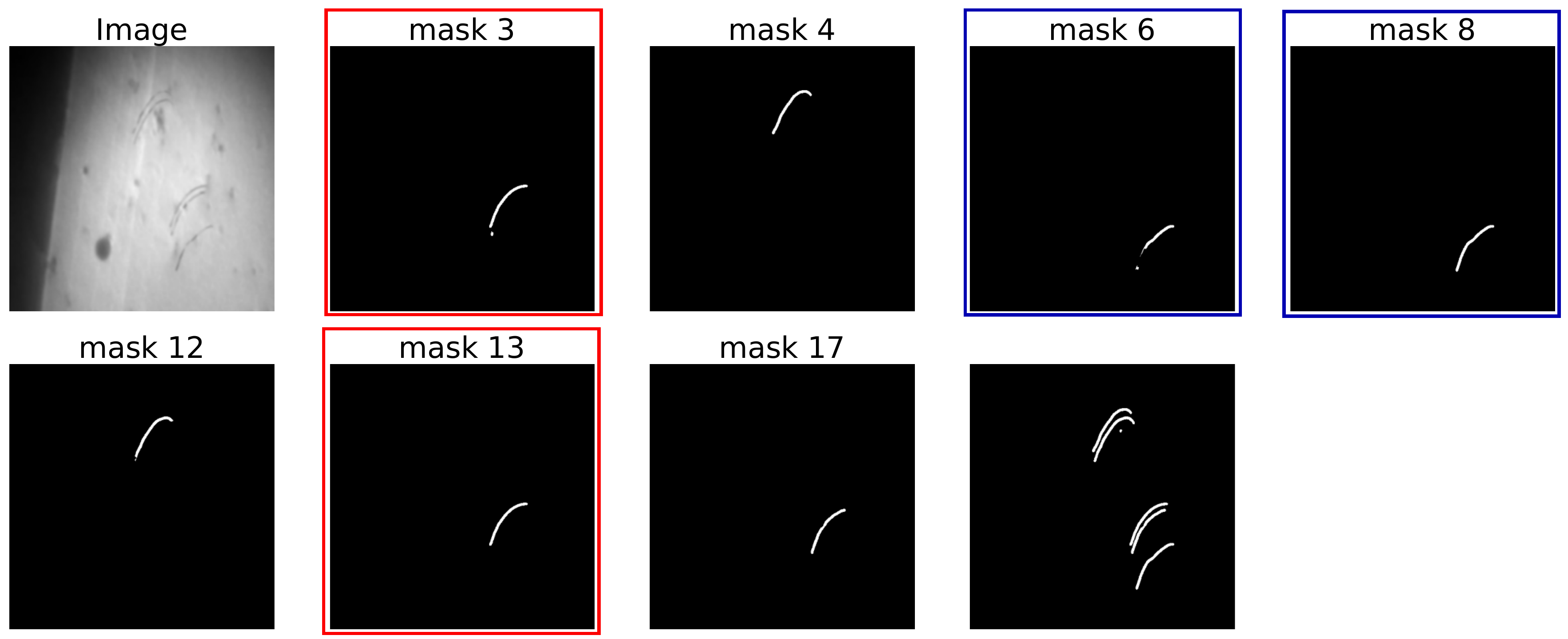}
	    \captionsetup{labelformat=empty}
		\caption*{Prediction for the synthetic twin of the real image}
	\end{minipage}
	\caption{Predictions on a real image from dataset RD1 along with its synthetic twin}
	\label{fig:twin_predictions}
\end{figure}

We also show the predictions of the model on the real image as well as the synthetic twin in \cref{fig:twin_predictions}. We can see that predictions for the digital twin are much better compared to the real image. The machine learning method tries to predict each dislocation in only one mask which allows us to isolate any incorrect predictions as seen in the figure where the line which is incorrectly predicted as dislocation is in a separate mask than dislocations. This could be useful in post processing of the predictions to analyze the dislocation image data. The dislocations usually move 
along specific slip directions which can be used to find the dislocations that are part of the pileup. 

\bibliographystyle{plain}

\bibliography{sn-bibliography.bib}
\end{document}